\documentclass{article}

\PassOptionsToPackage{numbers, compress}{natbib}

\usepackage[preprint]{main}

\usepackage{microtype} 
\usepackage{graphicx}
\usepackage{booktabs} 

\usepackage{hyperref}

\usepackage{amsmath}
\usepackage{amsthm}
\usepackage{subfig}
\usepackage{amsfonts}
\usepackage{tikz}

\usetikzlibrary{matrix}

\makeatletter
\newtheorem*{rep@theorem}{\rep@title}
\newcommand{\newreptheorem}[2]{%
\newenvironment{rep#1}[1]{%
 \def\rep@title{#2 \ref{##1}}%
 \begin{rep@theorem}}%
 {\end{rep@theorem}}}
\makeatother

\makeatletter
\newtheorem*{rep@prop}{\rep@title}
\newcommand{\newprop}[2]{%
\newenvironment{rep#1}[1]{%
 \def\rep@title{#2 \ref{##1}}%
 \begin{rep@prop}}%
 {\end{rep@prop}}}
\makeatother

\newtheorem{prop}{Proposition}
\newtheorem{theorem}{Theorem}
\newtheorem{definition}{Definition}
\newtheorem{assumption}{Assumption}
\newreptheorem{theorem}{Theorem}
\newprop{prop}{Proposition}

\title{On a Built-in Conflict between Deep Learning and Systematic Generalization}

\author{%
Yuanpeng Li
\thanks{\texttt{yuanpeng16@gmail.com}}
}

\begin{document}

\maketitle

\begin{abstract}
In this paper, we hypothesize that internal function sharing is one of the reasons to weaken o.o.d. or systematic generalization in deep learning for classification tasks.
Under equivalent prediction, a model partitions an input space into multiple parts separated by boundaries.
The function sharing prefers to reuse boundaries, leading to fewer parts for new outputs, which conflicts with systematic generalization.
We show such phenomena in standard deep learning models, such as fully connected, convolutional, residual networks, LSTMs, and (Vision) Transformers.
We hope this study provides novel insights into systematic generalization and forms a basis for new research directions.
Source codes are available at \url{https://github.com/yuanpeng16/BCDLSG}.
\end{abstract}

\section{Introduction}

A fundamental property of artificial intelligence is generalizations, where a trained model appropriately addresses unseen test samples.
Many problems adopt the i.i.d. assumption.
On the other hand, in out-of-distribution (o.o.d.) or \textbf{systematic generalization}~\cite{lake2018generalization,mcclelland1987parallel}, test samples have zero probability in training distribution.
It is crucial for human learning and provides creation ability.
So machines are also encouraged to have such ability.

A stream of artificial intelligence has been Connectionism~\cite{feldman1982connectionist,rumelhart1986parallel}, which uses many simple neuron-like units richly interconnected and processed in parallel.
It was criticized that Connectionist models do not support systematic generalization well~\cite{fodor1988connectionism,marcus1998rethinking}.
\textbf{Deep learning}~\cite{lecun2015deep} originates from Connectionism, and various techniques have enabled multiple-layer modelings and improved performance on i.i.d. problems in recent years.
Also, specific algorithms have been proposed to equip deep learning with systematic generalization ability~\cite{russin2019compositional,lake2019compositional}.
However, less discussion has been made on why standard deep learning models do not achieve systematic generalization.

This paper addresses the above question by looking into a build-in characteristic of deep learning.
Rich connections allow nodes on a layer to share inputs from nodes on the previous layer.
Each node is a function of the network input, so we call it function sharing (not parameter sharing).
We hypothesize that function sharing is one of the reasons to prevent systematic generalization in deep learning.
Under equivalent prediction, a classification network partitions an input space into multiple parts separated by boundaries.
Function sharing prefers to reuse boundaries and avoid redundant ones for train predictions.
It leads to fewer parts for new outputs and weakens systematic generalization.
The nearest neighbor classifier is an analogy for this effect because it predicts a training sample output for any test sample, so it does not have any part for new outputs.
We also discuss that function sharing belongs naturally to deep learning, so it may not be removed without additional design.

Systematic generalization usually requires that a sample has multiple explanatory factors of variation~\cite{bengio2013representation}, and the generalization is enabled by producing an unseen combination of seen factor values.
For example, models trained on blue rectangles and green triangles predict blue triangles.
We adopt factors in designing experiments and developing intuitions.
It helps experiments because new outputs are only related to function sharing between factors (Section~\ref{sec:experiments}).
So we limit our claim to the cases for recombination of factors.

\begin{figure*}[!t]
\centering
\subfloat[
DL preferred model
]{
\includegraphics[width=0.23\textwidth]{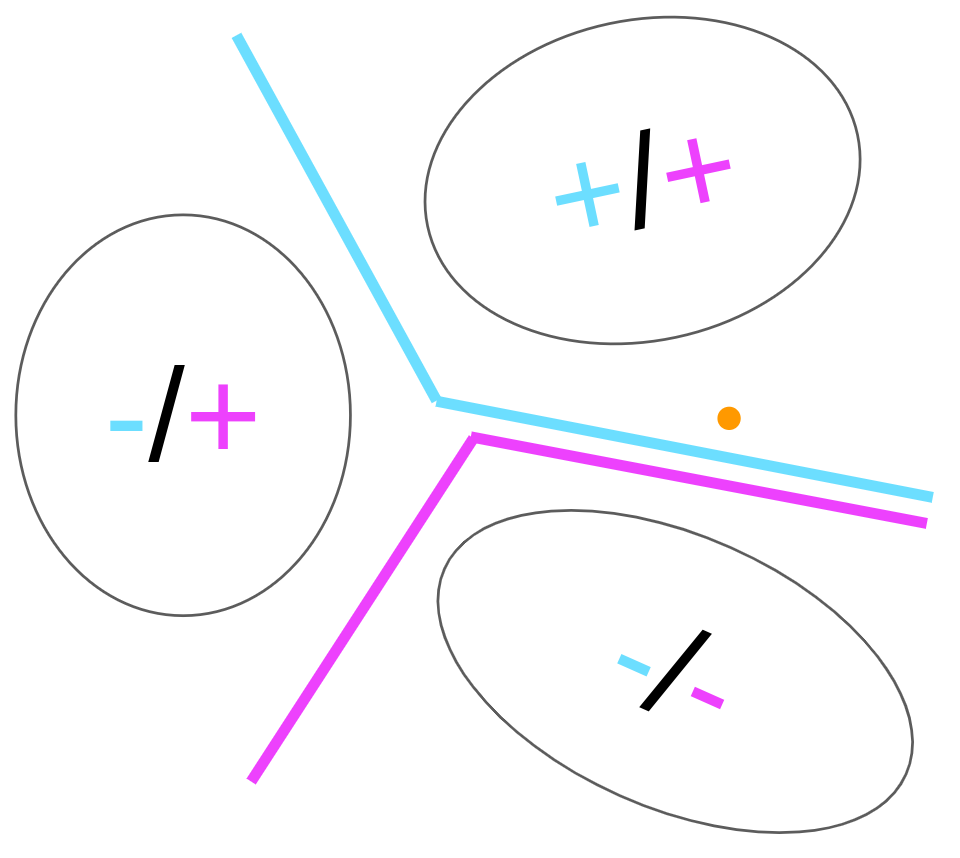}
\label{fig:conceptual_preferred}
}
\subfloat[
DL unpreferred model
]{
\includegraphics[width=0.23\textwidth]{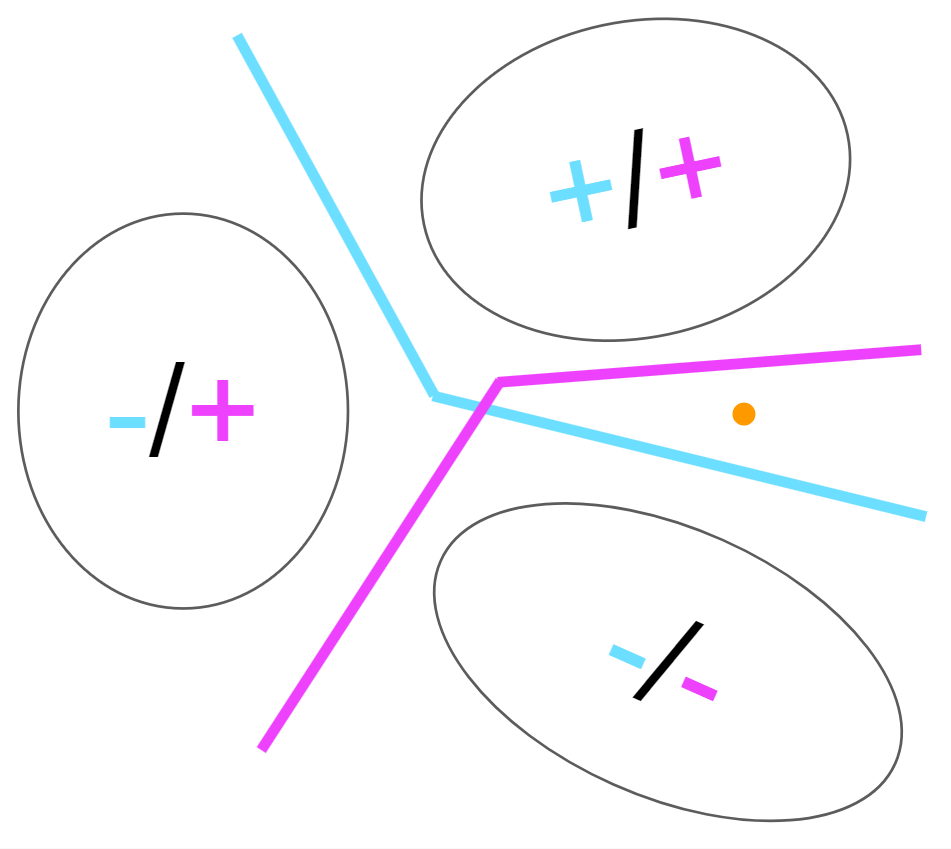}
\label{fig:conceptual_unpreferred}
}
\subfloat[
First function
]{
\includegraphics[width=0.23\textwidth]{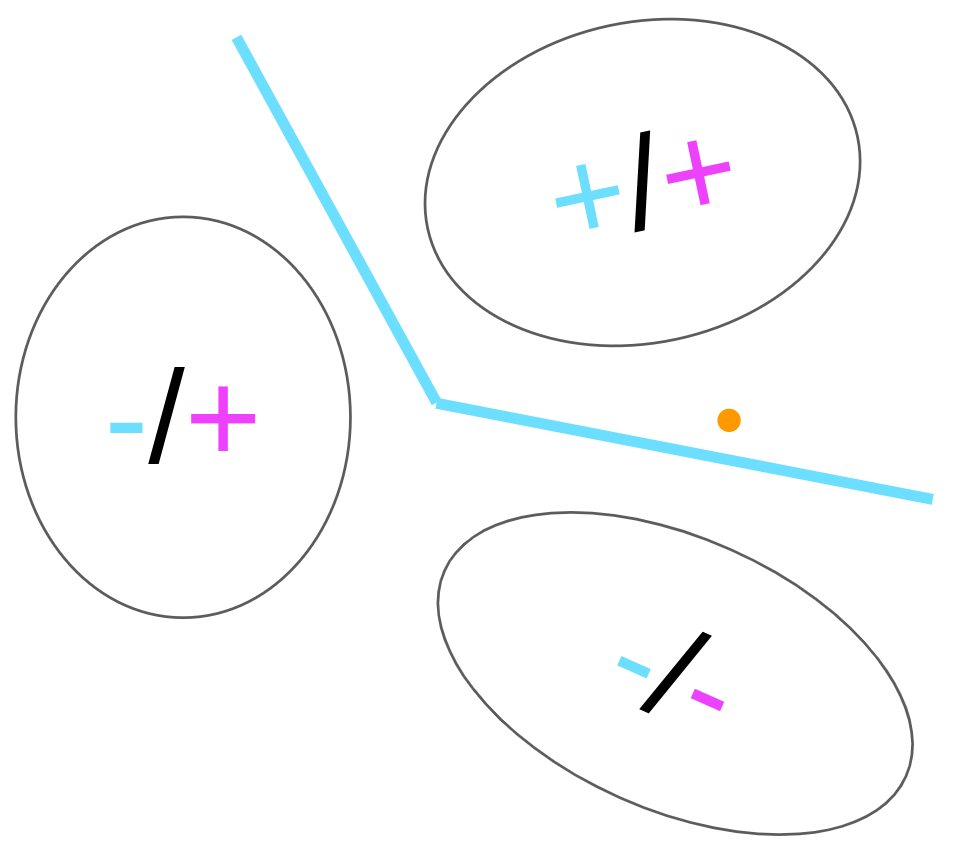}
\label{fig:first_function}
}
\subfloat[
Second function
]{
\includegraphics[width=0.23\textwidth]{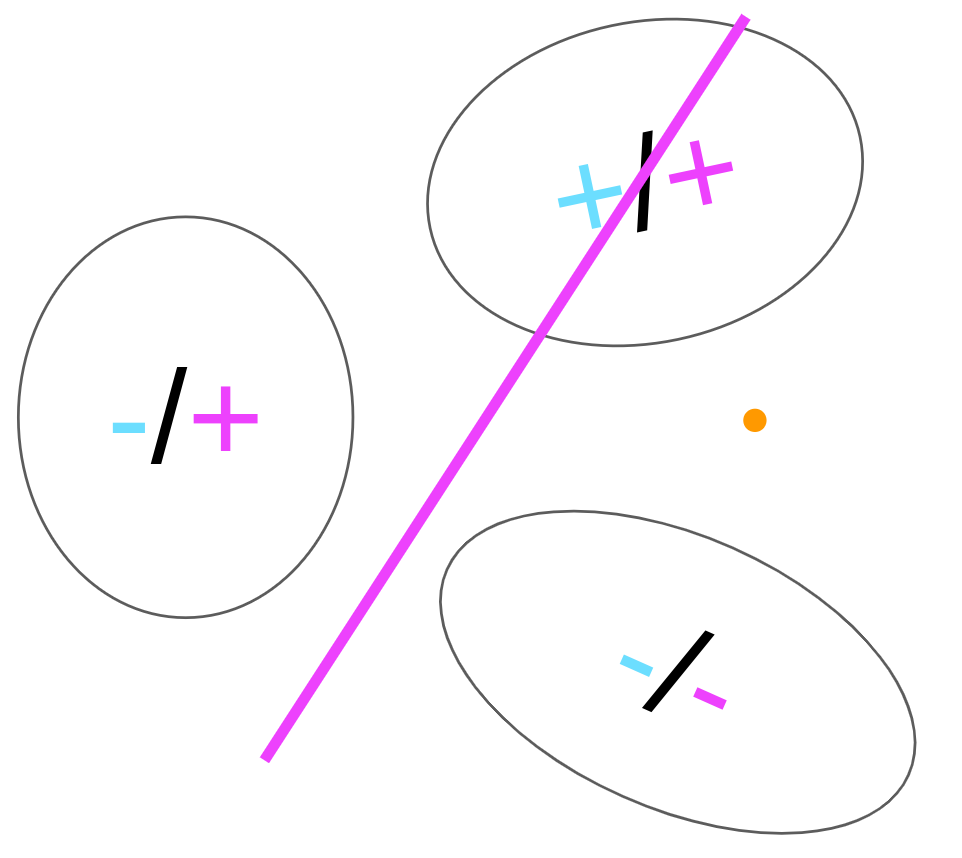}
\label{fig:second_function}
}
\caption{
Intuitions for the model preference of deep learning (DL).
Each figure is an input space containing three sets of training samples with outputs of +/+, -/-, and -/+, respectively.
Models have two decision boundaries (\textcolor{cyan}{cyan} and \textcolor{magenta}{pink}).
The orange dot is a test sample with a new ground-truth factor combination +/-.
We argue that the training process prefers the first case (a) over the second case (b) because it tends to share or reuse functions.
Suppose the first function (c) is learned, then the process likes to reuse it by learning a simple function (d) and combining them instead of learning the complicated pink function in (b) from scratch.
So the function between  +/+ and -/- is shared.
Few inputs are mapped to the new output, and systematic generalization is not achieved.
}
\label{fig:illustrations}
\end{figure*}

Figure~\ref{fig:illustrations} has an intuitive example.
A test sample (in orange) equals a set of training samples (+/+) on the first output.
Then they are also equal on the second output if the function is reused (see caption).
So they are equal in all the outputs, which conflicts with systematic generalization.
More generally, the two boundaries jointly partition an input space into multiple parts, and deep learning prefers (a) because it has fewer parts than (b).

This paper contributes to uncovering a built-in conflict between deep learning and systematic generalization.
We hope this study provides novel insights, forms a basis for new research directions, and helps improve machine intelligence to the human level.

\section{Related Work}
We look at the related work and development of systematic generalization and deep learning.
We then discuss recent research directions.

\paragraph{Systematic generalization and deep learning}
Systematic generalization\footnote{It is also called compositional generalization in other literature.} is considered the ``Great Move'' of evolution, caused by the need to process an increasing amount and diversity of environmental information~\cite{10.5555/86564}.
Cognitive scientists see it as central for an organism to view the world~\cite{gallistel2011memory}.
Studies indicate it is related to the prefrontal cortex~\cite{1995Relational}.
It was discussed that commonsense is critical~\cite{1959Program,1986CYC} for systematic generalization, and recent works aim to find general prior knowledge~\cite{goyal2020inductive}, e.g., Consciousness Prior~\cite{bengio2017consciousness}.
Levels of systematicity were defined~\cite{hadley1992compositionality,Niklasson94canconnectionist}, and types of tests were summarized~\cite{ijcai2020-708}.
We focus on the primary case with an unseen combination of seen factor values.

A closely related field is causal learning, rooted in the eighteenth-century~\cite{hume2003treatise} and classical fields of AI~\cite{pearl2003causality}.
It was mainly explored from statistical perspectives~\cite{pearl2009causality,peters2016causal,greenland1999causal,pearl2018does} with do-calculus~\cite{pearl1995causal,pearl2009causality} and interventions~\cite{peters2016causal}.
The causation forms Independent Causal Mechanisms (ICMs)~\cite{peters2017elements,scholkopf2021toward}.
Systematic generalization is the counterfactual when the joint input distribution is intervened to have new values with zero probability in training (covariate shift).
This work indicates that standard neural networks do not prefer to learn ICMs, i.e., they do not infer but learn associations~\cite{hinton1990mapping}.

Parallel Distributed Processing (PDP) models~\cite{rumelhart1986parallel} use Connectionist models with distributed representations, which describe an object in terms of a set of factors.
Though they have the potential to combine the factors to create unseen object representations~\cite{hinton1990mapping}, it was criticized that they do not address systematic generalization in general~\cite{fodor1988connectionism,marcus1998rethinking}.
Deep learning is a recent PDP model with many achievements~\cite{lecun2015deep, he2016resnet}.
It was studied that deep neural networks use the composition of functions to achieve high performance~\cite{NIPS2014_109d2dd3}.
The improvements in i.i.d. problems encourage to equip deep learning with systematic generalization.

\paragraph{Recent directions}
Other than architecture design~\cite{russin2019compositional,andreas2016neural} and data augmentation~\cite{andreas-2020-good,akyurek2020learning}, the main perspectives for systematic generalization approaches include disentangled representation learning, attention mechanism, and meta-learning.

Disentangled representation~\cite{bengio2013representation} is learned in unsupervised manners.
Early methods learn the representation from statistical independence~\cite{higgins2017beta,locatello2019challenging}.
Later, the definition of disentangled representation was proposed with symmetry transformation~\cite{higgins2018towards}.
It leads to Symmetry-based Disentangled Representation Learning~\cite{NEURIPS2019_36e729ec,NEURIPS2020_e449b931,NEURIPS2020_9a02387b,NEURIPS2020_c9f029a6}.
A disentangled representation learning model can be used as a feature extractor for other systematic generalization tasks.

Attention mechanisms are widely used in neural networks~\cite{bahdanau2015neural}.
Transformers~\cite{vaswani2017attention} are modern neural network architectures with self-attention.
Recurrent Independent Mechanisms~\cite{goyal2019recurrent} use attention and the name of the incoming nodes for variable binding.
Global workspace~\cite{goyal2021coordination} improves them by using limited-capacity global communication to enable the exchangeability of knowledge.
Discrete-valued communication bottleneck~\cite{liu2021discrete} further enhances systematic generalization ability.

Meta-learning~\cite{lake2019compositional} usually designs a series of training tasks for learning a meta-learner and uses it in a target task.
Each task has training and test data, where test data requires systematic generalization from training data.
When ICMs are available, they can be used to generate meta-learning tasks~\cite{scholkopf2021toward}.
Meta-reinforcement learning was used for causal reasoning~\cite{dasgupta2019causal}.
Meta-learning can also capture the adaptation speed to discover causal relations~\cite{bengio2020a,ke2019learning}.

Deep learning is a fast-growing field, and many efforts focus on designing architectures and algorithms to improve its performance.
However, it is less discussed why standard deep learning models do not achieve systematic generalization.
We look into a build-in conflict.

\section{A Built-in Conflict}
\label{sec:builtin_conflict}
We hypothesize a built-in conflict between the function sharing in deep learning and systematic generalization.
We first consider the requirements from visualized examples.
We then cover the definition, assumptions, and derivations of the conflict.

\subsection{Visualized examples and requirements}
\begin{figure*}[!ht]
\captionsetup[subfigure]{labelformat=empty}
\centering
\subfloat[
(a) First output
]{
\includegraphics[width=0.15\textwidth]{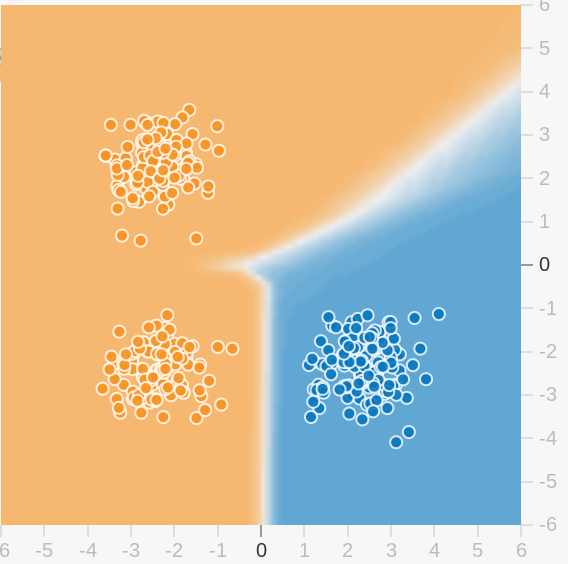}
\label{fig:linear_first}
}
\subfloat[
Second output
]{
\includegraphics[width=0.15\textwidth]{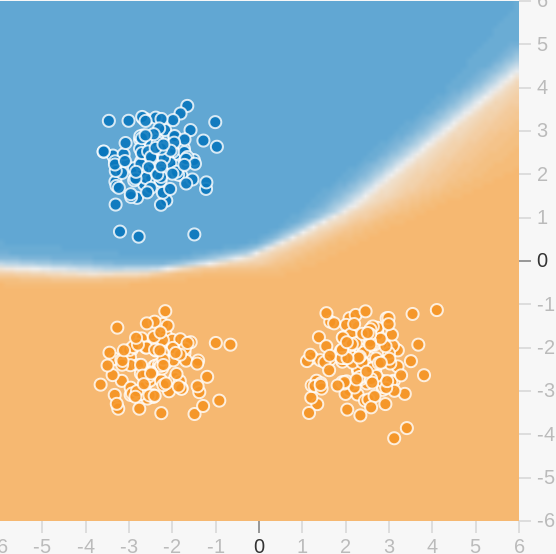}
\label{fig:linear_second}
}
\subfloat[
Result
]{
\includegraphics[width=0.15\textwidth]{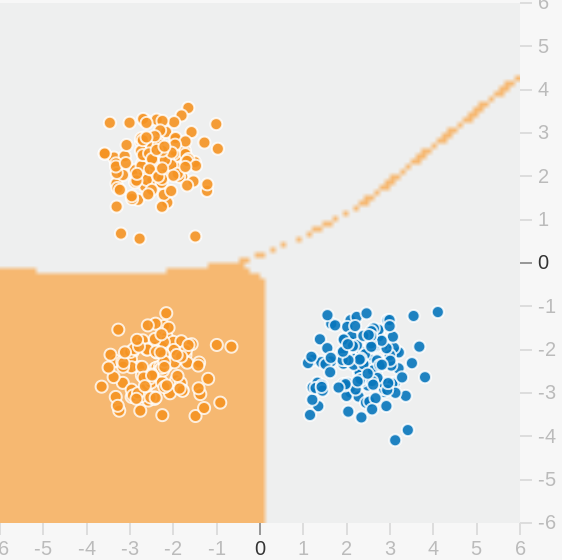}
\label{fig:linear_both}
}
\quad
\subfloat[
(b) First output
]{
\includegraphics[width=0.15\textwidth]{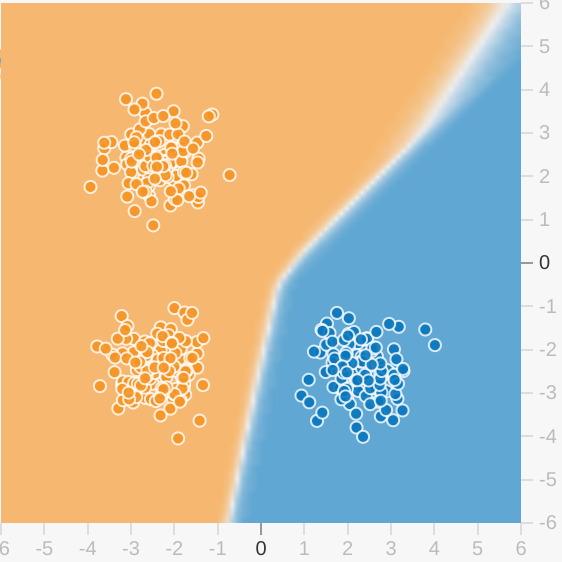}
\label{fig:case2_output1}
}
\subfloat[
Second output
]{
\includegraphics[width=0.15\textwidth]{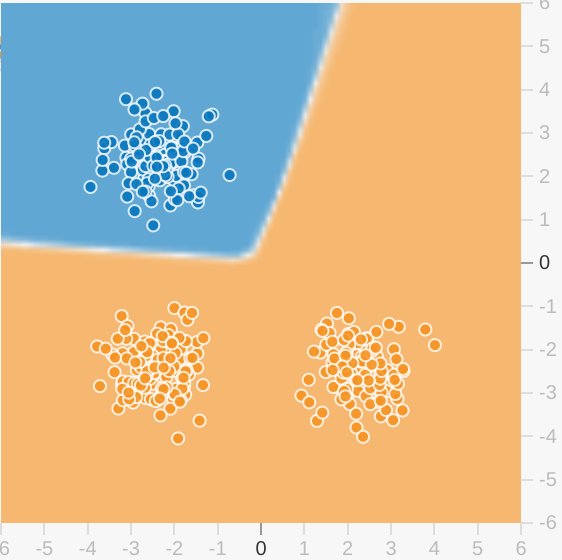}
\label{fig:case2_output2}
}
\subfloat[
Result
]{
\includegraphics[width=0.145\textwidth]{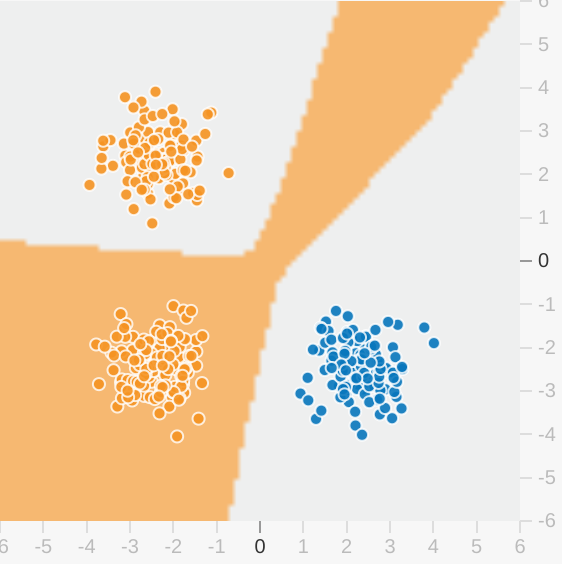}
\label{fig:case2_boundary}
}
\\
\subfloat[
(c) First output
]{
\includegraphics[width=0.15\textwidth]{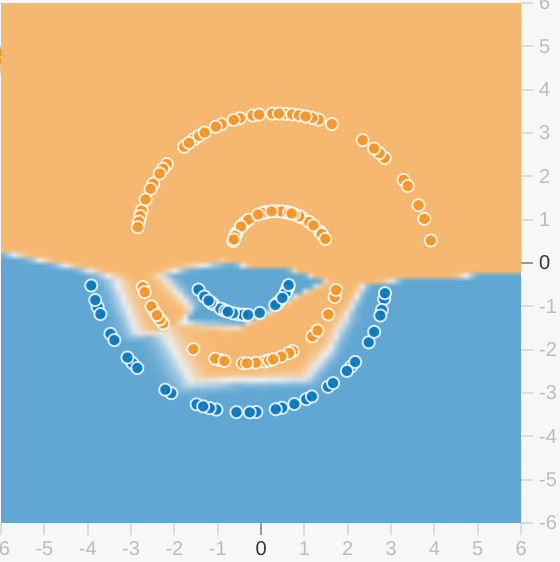}
\label{fig:case4_output1}
}
\subfloat[
Second output
]{
\includegraphics[width=0.15\textwidth]{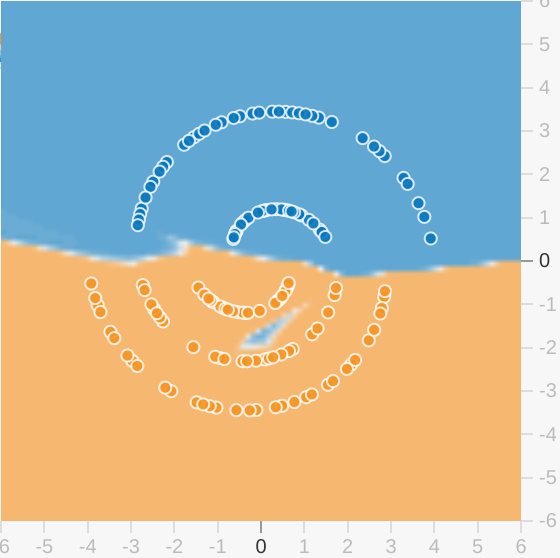}
\label{fig:case4_output2}
}
\subfloat[
Result
]{
\includegraphics[width=0.145\textwidth]{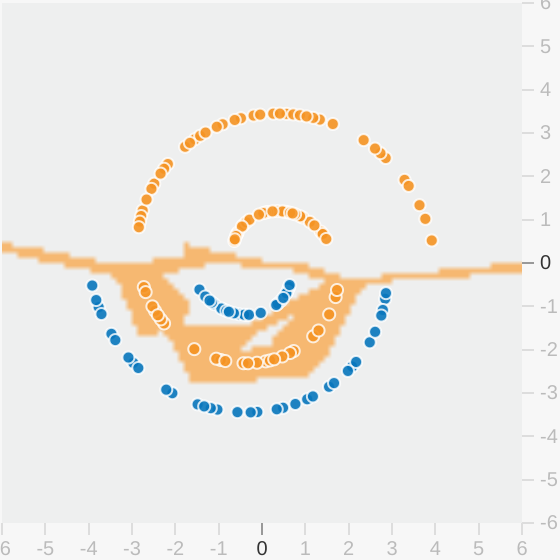}
\label{fig:case4_boundary}
}
\quad
\subfloat[
(d) First output
]{
\includegraphics[width=0.15\textwidth]{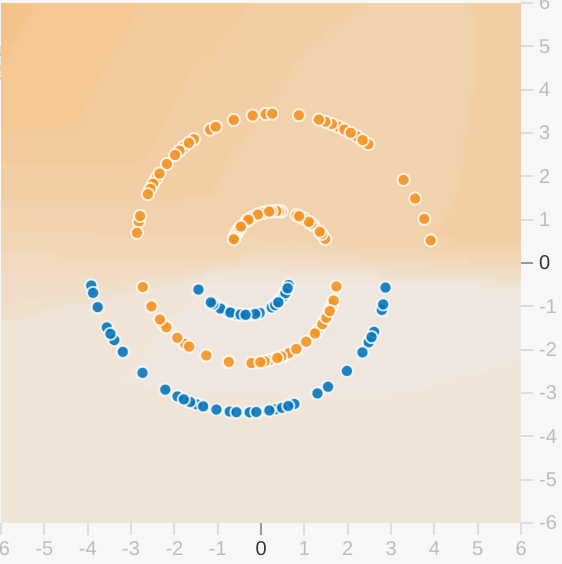}
\label{fig:case7_output1}
}
\subfloat[
Second output
]{
\includegraphics[width=0.15\textwidth]{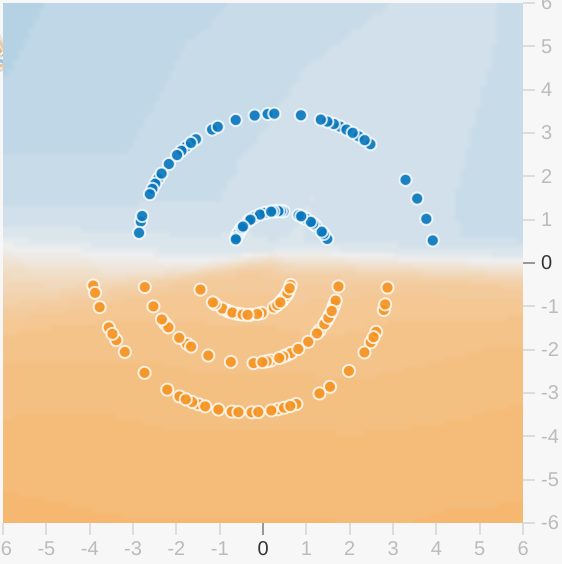}
\label{fig:case7_output2}
}
\subfloat[
Result
]{
\includegraphics[width=0.145\textwidth]{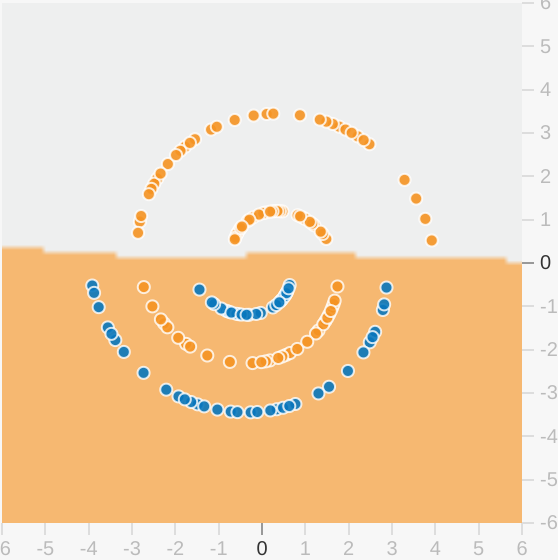}
\label{fig:case7_boundary}
}
\caption{
Visualized examples with a two-dimensional input and two binary outputs.
The model is a multiple-layer fully connected neural network.
Each set of three figures contains the predictions for the first and the second outputs and classification results for a model.
Each figure is the input space.
The dots are training samples whose colors are ground-truth labels.
For output figures, the background colors are model predictions.
For result figures, the background is orange if the prediction equals the new combination in no output, white if one, and blue if both.
The result figures do not have a blue background, so the models do not predict new label combinations for any input.
Hence systematic generalization is disabled.
Cases (a), (b), and (c) are trained models with correct training predictions.
Case (d) is after one training step, and the training prediction is not yet correct.
}
\label{fig:explanatory_examples}
\end{figure*}

Figure~\ref{fig:explanatory_examples} has visualized examples.
Case (a) is similar to the illustration example (Figure~\ref{fig:illustrations}).
The two functions are shared in the top-right region.

We may infer that function sharing prefers to merge two regions if it keeps the training loss.
In case (b) result figure, the top-right region is merged with the training region on the bottom-left.

In case (c) second output, the blue training region in the upper half and a small blue region in the middle-lower part have the same value in both the first and the second outputs.
They are disconnected in the input space and may be adjacent in a hidden layer.

Case (d) shows that new output combinations are avoided before the model is trained well.

The above cases encourage assumptions independent of the distribution geometry, and we should also consider the training process.
Note that (a) (b) and (c) (d) have different data distributions to show diverse phenomena.
Please refer to Appendix~\ref{sec:binary_classification_detail} for settings and Appendix~\ref{sec:more_visualization_example} for more examples.

\subsection{Systematic generalization}
We have input set $X$ and output set $Y$.
$Y$ contains $K$ factors $Y_1,\dots,Y_K$, which may be entangled.
The training set $\mathcal{D}_\text{train}$ has input $X_\text{train}$ and output $Y_\text{train}$.
The test set $\mathcal{D}_\text{test}$ has input $X_\text{test}$ and output $Y_\text{test}$.
In systematic generalization, $Y_\text{train}$ and $Y_\text{test}$ are disjoint.
However, the values for each factor $i$ are included in the training output.
A model $f$ maps input $X$ to the prediction of output $f(X)$.
A model enables systematic generalization if it correctly predicts the ground-truth test outputs.
\begin{definition}[Systematic generalization]
\label{definition:systematic_generalization}
The dataset requires that any test label is not a training label, but each factor of a test label is seen in a training label.
\begin{align*}
    \forall (x, y) \in \mathcal{D}_\text{test}: y \not\in Y_\text{train}, \quad \forall i = 1, \dots,  K, \exists y' \in Y_\text{train}: y'_i = y_i
\end{align*}
A function $f$ enables systematic generalization if $\forall (x, y) \in \mathcal{D}_\text{test}: y = f(x)$.
\end{definition}
When the model is well trained, we assume it correctly predicts training samples.
\begin{assumption}[Correct training prediction]
\label{assumption:correct_traininig_prediction}
$\forall (x, y) \in \mathcal{D}_\text{train}: y = f(x)$.
\end{assumption}

\subsection{Function sharing}
We assume a property of function sharing (intuition in Figure~\ref{fig:illustrations}).
Function sharing prefers to merge two regions to reuse functions and avoid learning redundant functions with the same effect on training data.
It means, for models $f$ and $g$, deep learning more or equally prefers $f$ over $g$ if the partition under $g$ is a refinement of that under $f$.
Equivalently, if two inputs have an identical prediction in $g$, their predictions are still equal in $f$.
It means an equal prediction in $g$ implies that in $f$.
\begin{assumption}[Function sharing]
\label{assumption:function_sharing}
Deep learning more or equally prefers $f$ over $g$ if
\begin{align*}
    \forall x_a, x_b \in X: g(x_a) = g(x_b) \implies f(x_a) = f(x_b)
\end{align*}
\end{assumption}
It is a bias in the learning process other than training loss.
Note that the function sharing does not depend on the geometry of distributions.

We consider what causes the function sharing.
Intuitively, the preference comes from greedy optimization and the task requirement to learn a complicated model.
A model learns to split or partition inputs by different predictions.
Some splits (can be a factor) might be easier to learn than others.
For example, learning to split inputs by color is more straightforward than by shape.
Then the greedy optimization learns it quickly and reuses its intermediate functions to learn other splits.
It is not likely that multiple splits (or factors) are learned equally fast during the whole training process.

In analogy, the mechanism of reusing function is similar to auxiliary tasks.
The parameters are updated to address a complicated main task while quickly learning and keeping the prediction ability for more straightforward auxiliary tasks.
A more frequent but less similar example is pre-training.
Pre-trained modules, such as image feature extractors or word embeddings, share the internal functions with the main target tasks.
However, the pre-training and the main task training do not happen simultaneously.
We will discuss more insights in Section~\ref{sec:discussion}.

\subsection{The conflict}
We derive propositions for proving the theorem and explaining the reasons for the phenomena.
In practice, the assumptions may not hold strictly, and the effects come more from the soft biases of the conclusions.
The proofs are in Appendix~\ref{sec:proofs}.

Assumption~\ref{assumption:function_sharing} leads a model to predict training outputs for any input.
\begin{prop}[Seen prediction]
\label{prop:seen_prediction}
From Assumption~\ref{assumption:function_sharing}, $\forall x \in X: f(x) \in f(X_\text{train})$.
\end{prop}
Informally, if $f(x) \not\in f(X_\text{train})$, the function $f$ has to differentiate this input $x$ from training inputs.
However, we can design another function $f'$ that predicts a training output for the input and keeps predictions for other inputs.
Then both $f$ and $f'$ equivalently distinguish training outputs.
With Assumption~\ref{assumption:function_sharing}, $f'$ is preferred, hence $\forall x \in X: f(x) \in f(X_\text{train})$.
It does not apply Assumption~\ref{assumption:correct_traininig_prediction}, indicating that the phenomena may happen before a model is well trained (Figure~\ref{fig:explanatory_examples}d).

If a model works well for the training set, it predicts training ground-truth output for any input.
\begin{prop}[Seen label]
\label{prop:seen_label}
From Assumption~\ref{assumption:correct_traininig_prediction} and Proposition~\ref{prop:seen_prediction}, $\forall x \in X: f(x) \in Y_\text{train}$.
\end{prop}
It says that the prediction is not any new output.
It is a stronger argument than avoiding a particular new output for each input in systematic generalization.
It explains that prediction is incorrect because any new output is resisted.
We evaluate it in Section~\ref{sec:experiments}.

We then look at the conflict. For any test data $(x, y)$, the definition of systematic generalization requires that the output $y$ should not be a training output $Y_\text{train}$.
However, this contradicts Proposition~\ref{prop:seen_label}.
\begin{theorem}[The conflict]
\label{theorem:the_conflict}
From Definition~\ref{definition:systematic_generalization} and Proposition~\ref{prop:seen_label}, $\forall (x, y) \in \mathcal{D}_\text{test}: y \neq f(x)$.
\end{theorem}

We covered systematic generalization definition and function sharing assumption.
We then derived propositions and a theorem of the conflict.

\section{Experiments}
\label{sec:experiments}
We run experiments to show that function sharing reduces the ability of systematic generalization in deep learning.
%
We not only compare sharing or not but also adjust the degree of sharing.
We focus on the cases where new outputs are unseen combinations of seen factor values.
We cover different standard deep neural network models.
The details of networks and experiments can be found in Appendix~\ref{sec:experiment_settings}.
The results of zero-shot learning datasets~\cite{farhadi2009describing,xian2019zero,wah2011the,patterson2012sun} are in Appendix~\ref{sec:zeroshot}, where factors are mainly related to the input locality~\cite{sylvain2020locality}.
We look at the experiment settings and results.

\subsection{Settings}
\paragraph{Data preparation}
We construct a dataset from two ten-class classification datasets.
The training data are generated from the original training dataset.
$Y_1$ is chosen from all possible labels.
$Y_2$ is chosen from five classes $\{Y_1, Y_1+1, \dots, Y_1+4\}$ (we use modular for labels).
The test data are generated from the original test dataset.
$Y_1$ is chosen in the same way as in training.
$Y_2$ is chosen from the other classes $\{Y_1+5, Y_1+6, \dots, Y_1+9\}$.
In this design, training and test label combinations are mutually exclusive, but test labels for each output factor are seen in training.
Any factor label appears evenly in training and test combinations.
$X_1$ and $X_2$ are chosen conditioned on their labels $Y_1, Y_2$, respectively, and merged as the input $X$.
The original datasets and input merge methods vary for each experiment.
All the choices follow uniform distributions.

\paragraph{Model architecture}
To evaluate the influence of function sharing on new outputs, we like to change the function sharing ability while keeping other properties stable for a deep learning model.
So we modify function sharing related to new outputs.
Since a new output appears only as a new factor combination in our setting, we adjust the sharings between output factors.
It does not need to remove all sharings and avoid difficulties in model design.
We choose a layer and duplicate the following layers, keeping the number of all hidden nodes in each layer if feasible (Figure~\ref{fig:model_architecture}).
We call the former part a shared network and the latter parts individual networks.
Each individual network predicts one output factor, so the output is disentangled.
We will discuss entangled outputs in Section~\ref{sec:entangled_output}.
We keep the depth of the whole network and change the depth of the shared network and individual networks.
Note that only sharing the input layer indicates learning two separate models.

\tikzset{%
  block/.style    = {draw, rectangle, minimum height = 2.8em, minimum width = 1.2em},
}
\begin{figure}[!ht]
    \centering
    \begin{tikzpicture}[yscale=0.65]
    \path [every node]
      node [block] (layer1) at (0,1) {$x$}
      node [block] (layer2) at (1,1) {}
      node [block] (layer3) at (2,1) {$h$}
      node [block] (layer4A) at (3,2) {}
      node [block] (layer4B) at (3,0) {}
      node [block] (layer5A) at (4,2) {}
      node [block] (layer5B) at (4,0) {}
      node [block] (layer6A) at (5,2) {$\hat{y}_1$}
      node [block] (layer6B) at (5,0) {$\hat{y}_2$}
      ;
      \draw[-latex] (layer1) -- node {} (layer2) ;
      \draw[-latex] (layer2) -- node {} (layer3) ;
      \draw[-latex] (layer3) -- node {} (layer4A) ;
      \draw[-latex] (layer3) -- node {} (layer4B) ;
      \draw[-latex] (layer4A) -- node {} (layer5A) ;
      \draw[-latex] (layer4B) -- node {} (layer5B) ;
      \draw[-latex] (layer5A) -- node {} (layer6A) ;
      \draw[-latex] (layer5B) -- node {} (layer6B) ;
    \end{tikzpicture}
    \caption{
        Deep model architecture.
        $x$ is input, and $f(x)=\hat{y}_1, \hat{y}_2$ are outputs.
        We duplicate the layers after layer $h$.
        The shared network is up to $h$, and the individual networks are after $h$ (their depths are all two here).
        We change the shared network depth while maintaining the entire network depth so that the function sharing ability is adjusted while other network properties are stable.
    }
    \label{fig:model_architecture}
\end{figure}
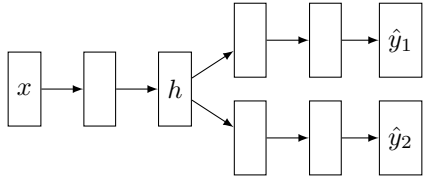

\paragraph{Evaluation metrics}
We use accuracy as the metric. A sample is correctly predicted if all the outputs are equal to the ground truth.
We have three types of accuracy.
The first is the regular evaluation of test data for systematic generalization (a: \textbf{Test Sample Accuracy}) corresponding to Theorem~\ref{theorem:the_conflict}.
We also consider a set of inputs mapped to unseen output combinations corresponding to Proposition~\ref{prop:seen_label}.
We evaluate whether the test samples predict one of the unseen output factor combinations (b: \textbf{Test Set Accuracy}).
However, test samples are only a subset of input space.
If a model learns a different set of factors, the expected inputs may not be those of test samples.
So we also evaluate systematic generalization as a model property for any valid input.
We randomly draw test inputs from the whole input space (c: \textbf{Random Set Accuracy})\footnote{$\delta(\cdot)$ is 1 if the statement is true and 0 otherwise. $U[\mathcal{X}]$ is the uniform distribution of valid inputs.}.
We run each experiment five times and plot the mean and the standard deviation (Figure~\ref{fig:diagonal_results}).
Please also refer to the result numbers in Table~\ref{tab:accuracy} (Appendix~\ref{sec:experiment_results}).
\begin{align*}
    &  a:\ \mathbb{E}_{(x,y) \sim P(\mathcal{D}_\text{test})}[\delta(f(x) = y)]
    && b:\ \mathbb{E}_{x \sim P(X_\text{test})}[\delta(f(x) \in Y_\text{test})]
    && c:\ \mathbb{E}_{x \sim U[\mathcal{X}]}[\delta(f(x) \in Y_\text{test})]
\end{align*}

\subsection{Results}
\paragraph{Fully Connected Network}
We use an eight-layer fully connected neural network with a flattened image input.
We use the Fashion dataset~\cite{xiao2017online} and the MNIST dataset~\cite{lecun1998gradient}.
The datasets are uncomplicated to avoid the training data under-fitting for a fully connected neural network.
We merge the two inputs by averaging values at each input node.

\paragraph{Convolutional Network}
We use a convolutional neural network with six convolutional layers and two fully connected layers.
We use the CIFAR-10 dataset~\cite{Krizhevsky09learningmultiple} and the Fashion dataset~\cite{xiao2017online}.
We scale the input sizes to that of the larger one and broadcast gray images to colored ones.
We merge the inputs by averaging at each node.
We use the Fashion dataset as one factor because the average of two colored images can cause training data under-fitting for convolutional neural networks.

\paragraph{Residual Network}
We use ResNet50~\cite{7780459}, composed of five stages, and we treat each stage as a layer while changing the shared network depth.
It has the dataset setting in the CNN experiment.

\paragraph{Vision Transformer}
We use Vision Transformer~\cite{dosovitskiy2021an} with one fully connected layer for each patch, five attention layers, and two fully connected layers.
We treat the patches as one layer.
It has the dataset setting in the CNN experiment.

\paragraph{LSTM}
A recurrent network has the same parameters for each layer, so it does not support learning different individual networks.
Instead, we treat an LSTM as a layer.
We use stacked LSTM models with an embedding layer, five bidirectional LSTM layers, and two fully connected layers.
We use the Reuters dataset~\cite{10.1007/BFb0026683} for both the first and the second datasets.
We filter samples by a maximum input length of 200.
We use the most frequent ten classes of samples.
We merge inputs by concatenating two input texts.
Since the text lengths vary, the merged inputs have different lengths.

We also run experiments for one-layer LSTM, which compares sharing or not sharing all layers.
The results indicate that the shared network has less generalization than the individual network (Table~\ref{tab:accuracy}).

\paragraph{Transformer}
We use Transformer~\cite{vaswani2017attention}.
Since it is a classification problem, we only use the encoder.
It has one embedding layer, five hidden layers, and two fully connected layers.
We use the same dataset setting as the LSTM experiment.

\begin{figure*}[!ht]
\centering
\subfloat[
DNN
]{
\includegraphics[width=0.48\textwidth]{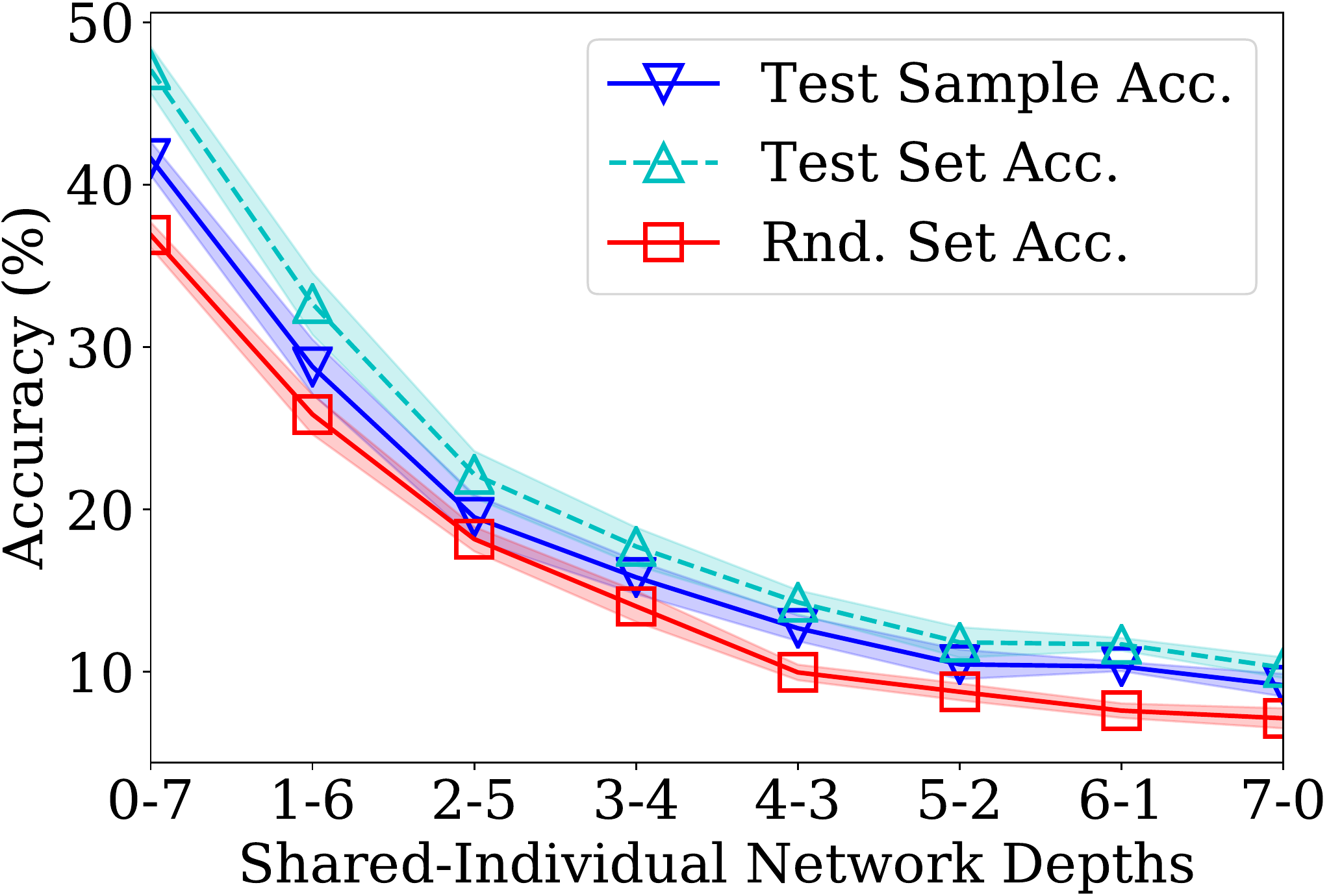}
\label{fig:dnn_diagonal}
}
\subfloat[
CNN
]{
\includegraphics[width=0.48\textwidth]{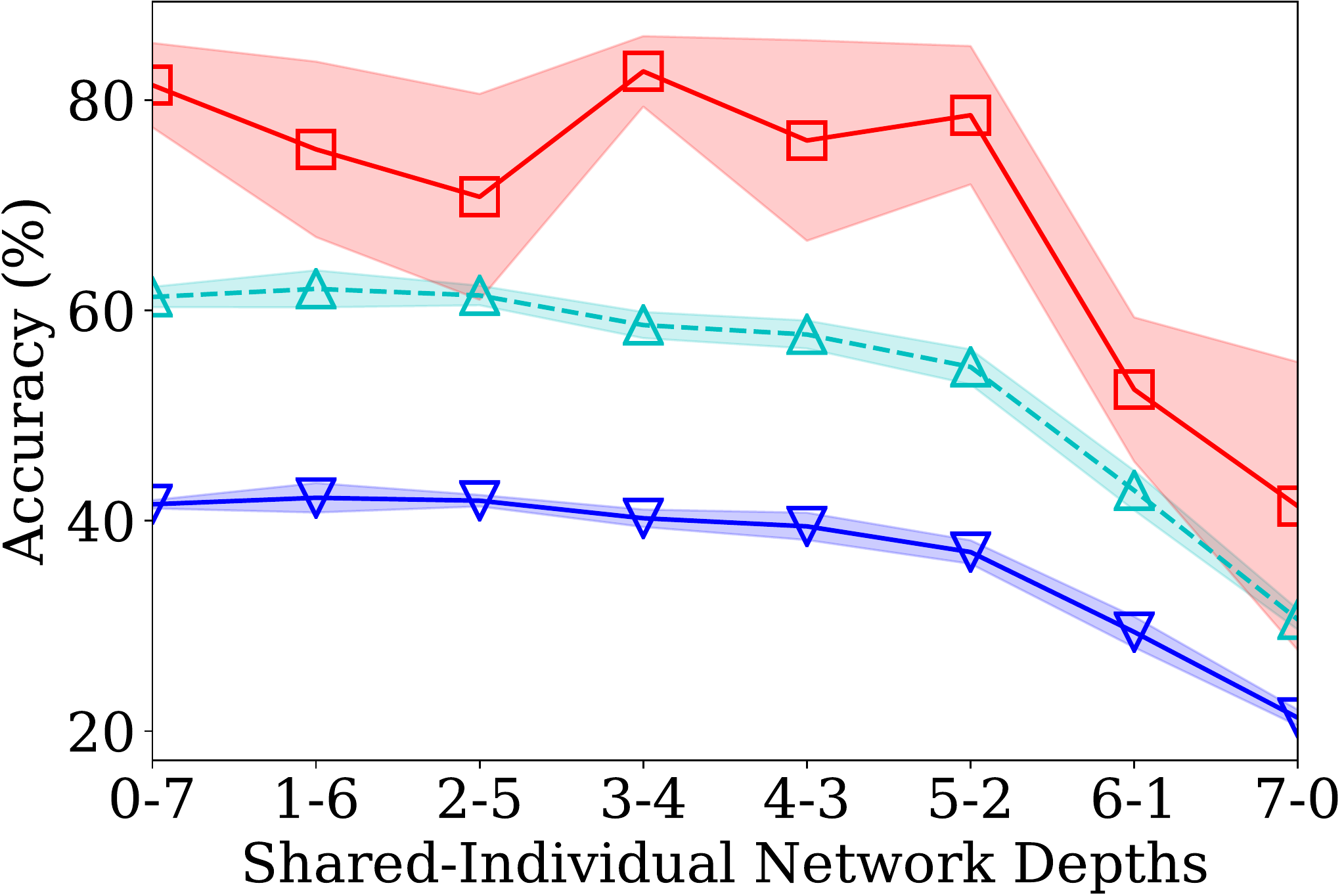}
\label{fig:cnn_diagonal}
}
\\
\subfloat[
ResNet
]{
\includegraphics[width=0.48\textwidth]{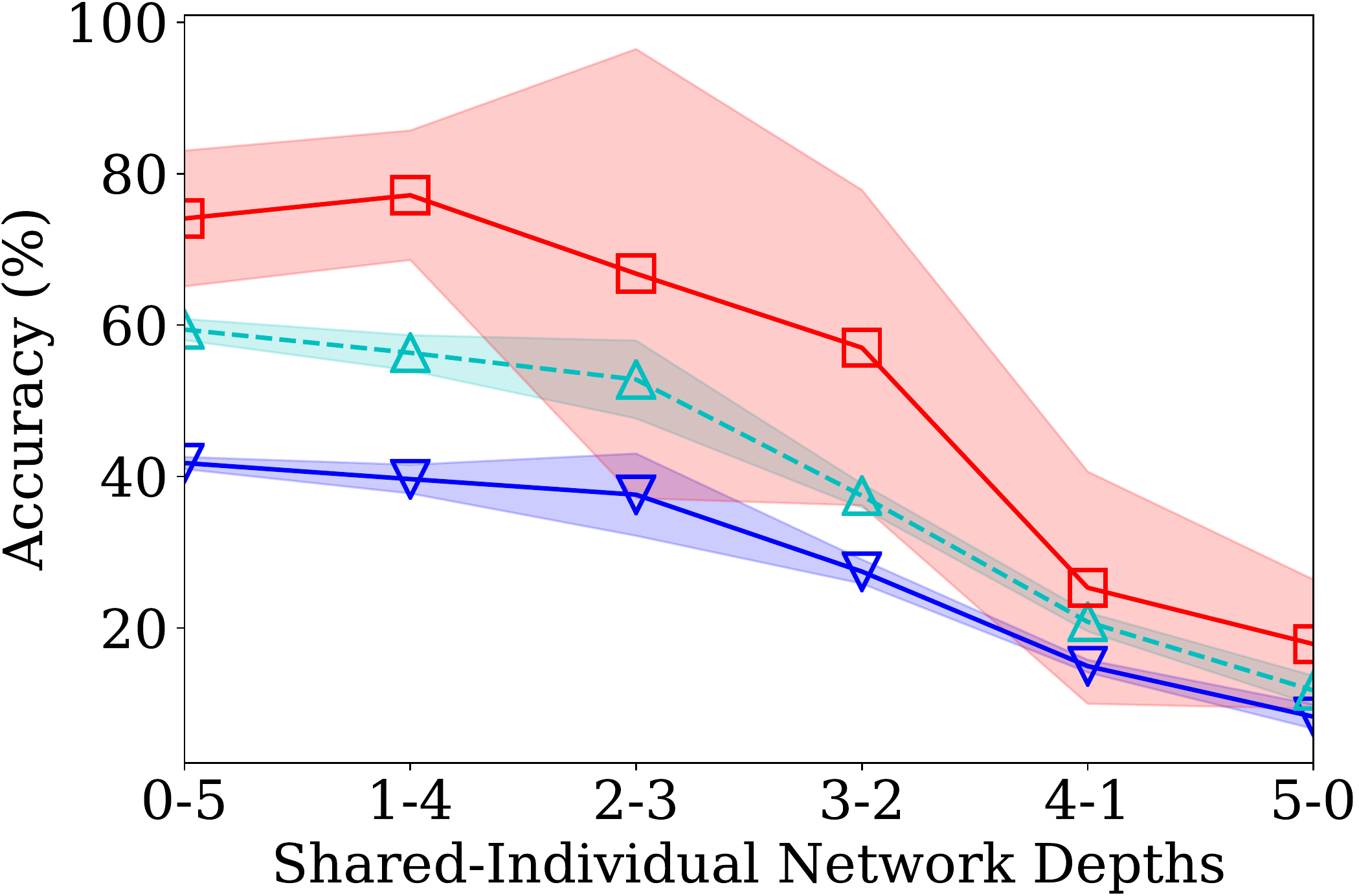}
\label{fig:resnet_diagonal}
}
\subfloat[
Vision Transformer
]{
\includegraphics[width=0.48\textwidth]{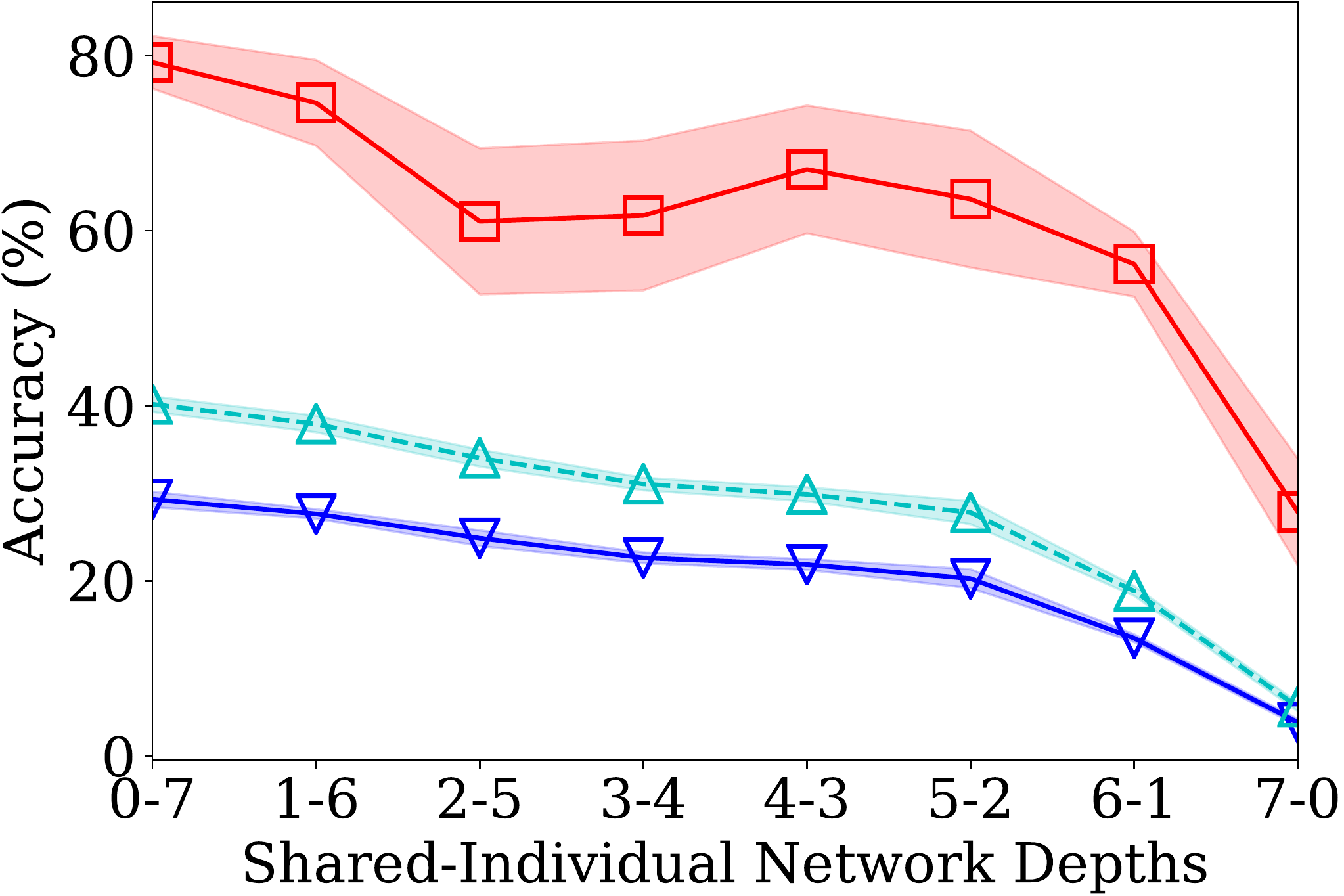}
\label{fig:cifar_fashion_added_diagonal_vit_acc}
}
\\
\subfloat[
LSTM
]{
\includegraphics[width=0.48\textwidth]{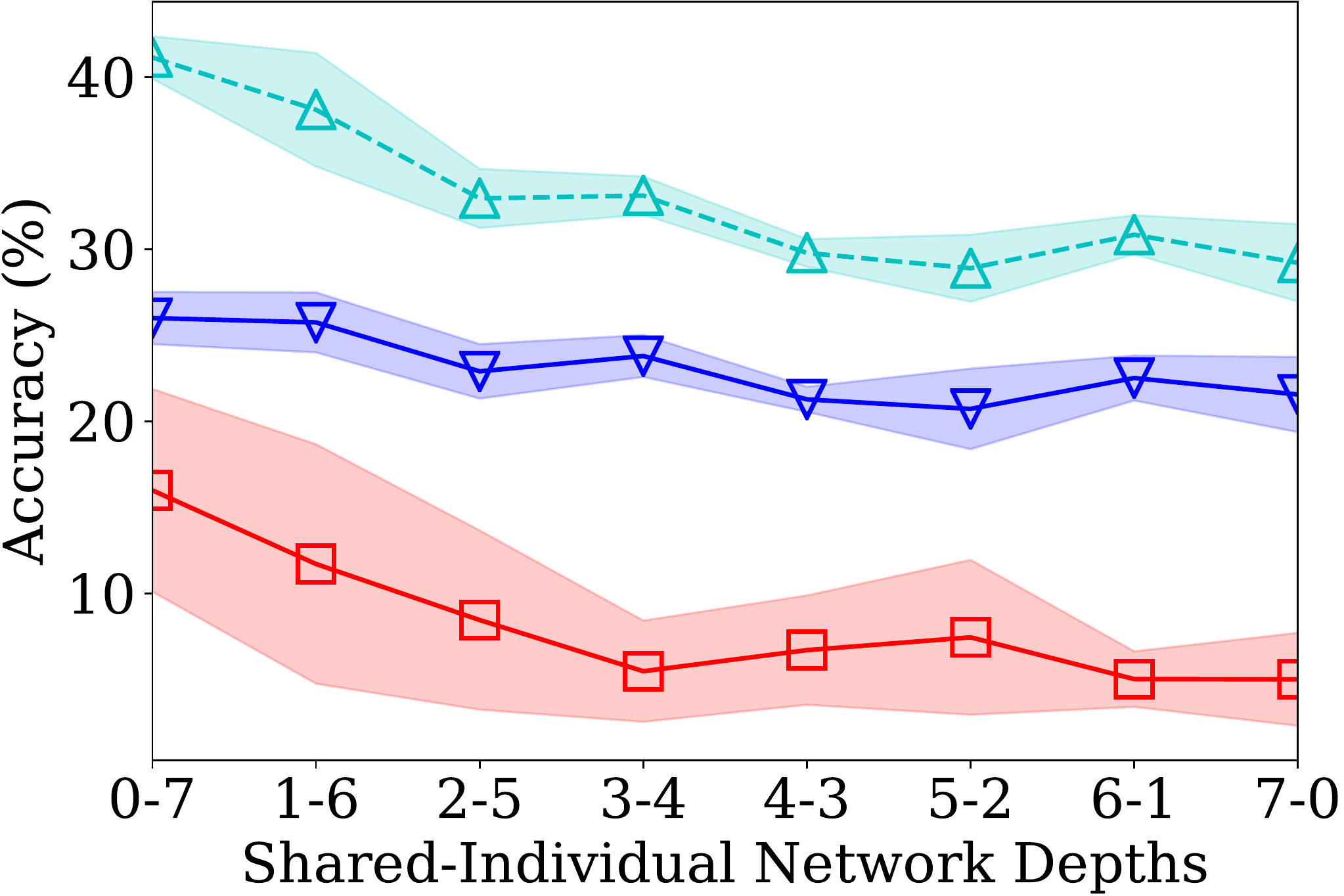}
\label{fig:lstm_diagonal}
}
\subfloat[
Transformer
]{
\includegraphics[width=0.48\textwidth]{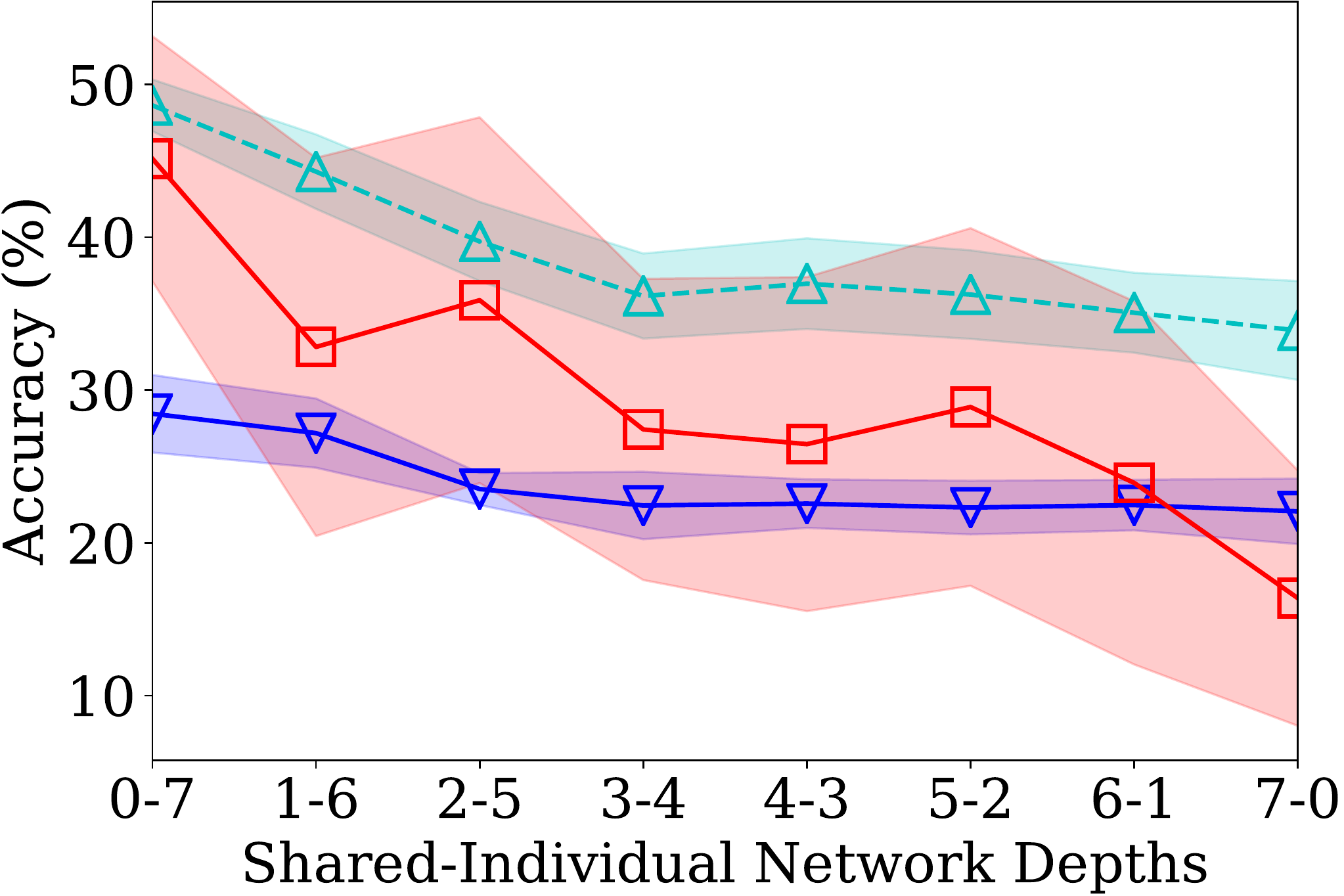}
\label{fig:transformer_diagonal}
}
\caption{
Results of three types of systematic generalization accuracy with shared and individual (sub)network depths.
As the shared network depth increases, the accuracy generally decreases, indicating that function sharing reduces systematic generalization.
Test Set Accuracy and Random Set Accuracy are the ratios of predicting unseen combinations.
Test Sample Accuracy and Test Set Accuracy are for the test data, and Random Set Accuracy is for randomly generated inputs.
}
\label{fig:diagonal_results}
\end{figure*}

\paragraph{Summary of results}
Figure~\ref{fig:diagonal_results} shows that, for each evaluation, the accuracy on the left end (not sharing any hidden layer) is higher than that on the right end (sharing all hidden layers), and it generally decreases as the shared network depth increases.
The results indicate that the function sharing weakens systematic generalization.

\section{Discussions}
\label{sec:discussion}

\subsection{Entangled output}
\label{sec:entangled_output}
Though the theory does not require disentangled output, the experiments use it to split the network into individual ones.
We discuss that entangled output prediction is not easier than disentangled one.
Hence the experiment conclusions will likely extend to entangled outputs.
(1) Disentangled output is a particular and usually less complicated case of entangled output.
(2) Entangled output can be seen as another shared layer, and the experiments show that increasing shared layers reduces systematic generalization ability, so entangled outputs are also likely to suffer from the problem.
(3) We can see each output node of an entangled output as a factor, so it becomes disentangled output.
If a value is not seen for an output node, it makes the generalization even more difficult.

\subsection{Beyond factor recombination}
The definition of systematic generalization (Definition~\ref{definition:systematic_generalization}) requires that each test label factor is seen in a training label.
However, it is not directly used in the derivations, and the conclusions may apply to more general o.o.d. problems other than recombining factors.
A new output may correspond to an unseen activation for an output node, e.g., a new class in a classification problem.
In such settings, it is sometimes discussed that the bias parameter of the output is a reason to avoid the new value prediction because it does not have any training signal to increase its value.
This work provides another reason: the whole network is not likely to predict a new value.

\subsection{Why does function sharing happen in deep learning?}
\label{sec:why_function_sharing}
We discuss that function sharing can be caused by characteristics of deep learning: deep architecture, shareable network, and greedy optimization.
Deep architecture and shareable networks make it possible for factors to share elaborated functions, and greedy optimization encourages the sharing.
Deep architecture and greedy search are necessary for deep learning.
Deep learning uses deep architecture to fit complicated non-linear functions.
Deep learning has a large and complex parameter space.
To search in it, we need some prioritization, which leads to a greedy search.
It might be gradient descent and its variants or other methods such as evolutionary computation.
The shareable network is widely used in standard deep learning models and works well for i.i.d. problems.
However, it is less critical compared to the other ones.

\subsection{Potential solutions}
We consider possible solutions to avoid function sharing and achieve systematic generalization.
From the above discussion, deep architecture and greedy optimization are essential for deep learning, so we look at shareable networks.
We consider the recombination of factors and focus on sharing between factors.
Then, one potential solution uses individual networks for output factors, similar to the experiment setup.
We discuss how to design networks when the input or the output is entangled.
If the output is entangled, we can design an architecture where each individual network only changes one factor in the output.
For example, one individual network changes color, and another changes shape.
If the input is entangled, we need to extract factors from it to feed the individual networks.
It contains two questions: how to avoid spurious influence from other factors and keep it working in test distribution.
We can bottleneck representations for the first one and divide the input into units invariant in training and test for the other, such as words or objects.

\section{Conclusions}
This paper investigates a built-in conflict between deep learning and systematic generalization.
It explains one of the reasons why standard neural networks seldom achieve systematic generalization.
We discuss that the conflict is caused by sharing of internal functions.
A model partitions an input space into multiple parts separated by boundaries.
The function sharing tends to reuse the boundaries, leading to fewer parts for new outputs, which conflicts with systematic generalization.
The experiments show the phenomena in different types of standard neural network models.
We hope this finding provides a new understanding of systematic generalization mechanisms and helps to improve machine learning algorithms for a higher level of artificial intelligence.

\clearpage
\acksection
We thank Liang Zhao for insightful discussions and beneficial suggestions. We also thank Kenneth Church and Yi Yang for providing valuable comments.

\bibliography{main}
\bibliographystyle{plain}
\clearpage
\appendix

\section{Proofs}
\label{sec:proofs}
\begin{repprop}{prop:seen_prediction}[Seen prediction]
From Assumption~\ref{assumption:function_sharing}, $\forall x \in X: f(x) \in f(X_\text{train})$.
\end{repprop}
\begin{proof}
Given a function $g$, we construct a function $f$.
We pick a $x' \in X_\text{train}$. $\forall x \in X$:
\begin{align*}
    g(x) \in g(X_\text{train}): \quad & f(x) = g(x), \\
    \text{o.w.}: \quad & f(x) = f(x').
\end{align*}
We look at the conditions. $\forall x_a, x_b \in X:$
\begin{align*}
    g(x_a) \in g(X_\text{train}): \quad & g(x_a) = g(x_b) \implies f(x_a) = g(x_a) = g(x_b) = f(x_b) \\
    \text{o.w.}: \quad & g(x_a) = g(x_b) \implies g(x_b) \not\in g(X_\text{train}) \implies f(x_a) = f(x') = f(x_b)
\end{align*}
On the other hand,
\begin{align*}
    \exists x \in X_\text{test}: g(x) \not\in g(X_\text{train}) \implies g(x) \neq g(x') \in g(X_\text{train}), \quad f(x) = f(x').
\end{align*}
So $f(x) = f(x')$ does not imply $g(x) = g(x')$.
With Assumption~\ref{assumption:function_sharing}, $f$ is preferred over $g$.
It follows that $\forall x \in X$:
\begin{align*}
    g(x) \in g(X_\text{train}): \quad & \exists x'' \in X_\text{train}: g(x) = g(x'') \implies f(x) = f(x'') \in f(X_\text{train}) \\
    \text{o.w.}: \quad & f(x) = f(x') \in f(X_\text{train})
\end{align*}
Therefore, $\forall x \in X: f(x) \in f(X_\text{train})$.
\end{proof}

\begin{repprop}{prop:seen_label}[Seen label]
From Assumption~\ref{assumption:correct_traininig_prediction} and Proposition~\ref{prop:seen_prediction}, $\forall x \in X: f(x) \in Y_\text{train}$.
\end{repprop}
\begin{proof}
From Assumption~\ref{assumption:correct_traininig_prediction}, $f(X_\text{train}) = Y_\text{train}$.
From Proposition~\ref{prop:seen_prediction},
\begin{align*}
    \forall x \in X: f(x) \in f(X_\text{train}) = Y_\text{train}
\end{align*}
\end{proof}

\begin{reptheorem}{theorem:the_conflict}[The conflict]
From Definition~\ref{definition:systematic_generalization} and Proposition~\ref{prop:seen_label}, $\forall (x, y) \in \mathcal{D}_\text{test}: y \neq f(x)$.
\end{reptheorem}
\begin{proof}
$\forall (x, y) \in \mathcal{D}_\text{test}$: $y \not\in Y_\text{train}$ (Definition~\ref{definition:systematic_generalization}), $x \in X_\text{test} \subseteq X \implies f(x) \in Y_\text{train}$ (Proposition~\ref{prop:seen_label}).\\
Therefore $y \neq f(x)$.
\end{proof}

\section{Experiment Details}
\label{sec:experiment_settings}

\subsection{Visualization settings}
\label{sec:binary_classification_detail}
The model is a fully connected neural network with two input and two output nodes.
It has six hidden layers with ReLU activations, and each hidden layer has eight nodes.
We use a mini-batch size of 10 with a learning rate of 0.01.
We iterate until the model prediction becomes stable.
Please see the original work for more information.
We use six Intel(R) Core(TM) i5-8400 2.80GHz CPUs, and the asset has a public license.

\subsection{Experiment settings}
\label{sec:main_experiments_detail}
We use GeForce GTX 1080 or GeForce GTX 1050 Ti GPU for single GPU experiments.
We use TensorFlow for implementation.
The assets have a public license.

Each input element is linearly scaled to [-0.5, 0.5] for image input.
We also uniformly sample from this interval for random image input.
We select two sentence lengths uniformly from valid integers (one to maximum length) and then generate each word uniformly from the vocabulary for random text input.

\paragraph{Fully Connected Network}
The input shape is 28 $\times$ 28, flattened to a vector.
There are seven fully connected layers.
Each of them has 512 hidden nodes and ReLU activation.
The output has ten nodes and Softmax activation.
%
We use cross-entropy loss and Adam optimizer with a learning rate of 0.001.
The batch size is 512, and we train 2,000 iterations.
Each evaluation uses 10,000 samples.

\paragraph{Convolutional Network}
The input shape is 32 $\times$ 32 $\times$ 3.
There are seven convolutional layers.
Each of them has 3 $\times$ 3 kernel size with 64 channels.
Then the layer is flattened.
We have a fully connected layer with 128 nodes and ReLU activation.
The output layer has ten nodes with Softmax activation.
We use cross-entropy loss and Adam optimizer with a learning rate of 0.001.
The batch size is 512, and we train 5,000 iterations.
Each evaluation uses 10,000 samples.

\paragraph{Residual Network}
The input is the same as CNN.
The model is the standard ResNet50 implementation.
The hidden groups are treated as one layer, so there are five hidden layers.
The hidden layer size is 64.
The output layer has ten nodes with Softmax activation.
We use cross-entropy loss and Adam optimizer with a learning rate of 0.001.
The batch size is 512, and we train 10,000 iterations.
Each evaluation uses 10,000 samples.

\paragraph{Vision Transformer}
The input is the same as CNN.
The model is the standard Vision Transformer implementation with seven hidden layers.
The hidden layer size is 64.
The output layer has ten nodes with Softmax activation.
We use cross-entropy loss and Adam optimizer with a learning rate of 0.001.
The batch size is 512, and we train 10,000 iterations.
Each evaluation uses 10,000 samples.

\paragraph{LSTM}
The vocabulary size is 30,977, including a start symbol and padding symbol.
The input length is 200.
The embedding size is 64.
There are seven stacked bidirectional LSTM layers, and each has 32 hidden nodes for each direction.
Then the output is flattened.
The output layer is a fully-connected layer with ten output nodes and Softmax activation.
We use cross-entropy loss and Adam optimizer with a learning rate of 0.001.
The batch size is 64, and we train 1,000 iterations.
Each evaluation uses 10,000 samples.

\paragraph{Transformer}
The input is the same as that of LSTM.
The embedding size is 64.
There are seven hidden groups.
The hidden layer size is 64.
The output is flattened.
The output layer is a fully-connected layer with ten output nodes and Softmax activation.
We use cross-entropy loss and Adam optimizer with a learning rate of 0.001.
The batch size is 64, and we train 2,000 iterations.
Each evaluation uses 10,000 samples.

\subsection{Experiment results}
\label{sec:experiment_results}
We numerically compare the individual (left ends in result figures, 0-max) and shared (right ends, max-0) networks in Table~\ref{tab:accuracy}.
LSTM is the stacked LSTM, and LSTM-1 has only one LSTM layer.
It shows that shared network has lower scores than individual network on the three types of accuracy.
So it indicates that sharing network avoids systematic generalization.

\begin{table}[!ht]
\centering
\caption{Accuracy (mean $\pm$ std \%). Experimental results.}
\label{tab:accuracy}
\begin{tabular}{lrr|rr|rr}
& \multicolumn{2}{c|}{Test Sample Accuracy}
& \multicolumn{2}{c|}{Test Set Accuracy}
& \multicolumn{2}{c}{Random Set Accuracy} \\
& \multicolumn{1}{c}{Individual} & \multicolumn{1}{c|}{Shared} 
& \multicolumn{1}{c}{Individual} & \multicolumn{1}{c|}{Shared}
& \multicolumn{1}{c}{Individual} & \multicolumn{1}{c}{Shared} \\
\hline
DNN & 41.6 {\small$\pm$ 1.0\par} & 9.2 {\small$\pm$ 0.7\par} & 47.1 {\small$\pm$ 1.4\par} & 10.2 {\small$\pm$ 0.6\par} & 36.9 {\small$\pm$ 0.8\par} & 7.1 {\small$\pm$ \ \ 0.6\par} \\
CNN & 41.6 {\small$\pm$ 0.4\par} & 21.3 {\small$\pm$ 0.8\par} & 61.3 {\small$\pm$ 1.0\par} & 30.6 {\small$\pm$ 1.0\par} & 81.4 {\small$\pm$ 4.0\par} & 41.4 {\small$\pm$ 13.7\par} \\
ResNet & 41.8 {\small$\pm$ 0.8\par} & 8.3 {\small$\pm$ 1.6\par} & 59.4 {\small$\pm$ 1.4\par} & 11.8 {\small$\pm$ 1.9\par} & 74.1 {\small$\pm$ 9.0\par} & 17.9 {\small$\pm$ \ \ 8.5\par} \\
ViT & 29.3 {\small$\pm$ 0.9\par} & 3.8 {\small$\pm$ 0.4\par} & 40.2 {\small$\pm$ 0.9\par} & 5.6 {\small$\pm$ 0.5\par} & 79.2 {\small$\pm$ 3.0\par} & 27.9 {\small$\pm$ \ \ 6.1\par} \\
LSTM & 26.0 {\small$\pm$ 1.5\par} & 21.6 {\small$\pm$ 2.2\par} & 41.1 {\small$\pm$ 1.2\par} & 29.2 {\small$\pm$ 2.3\par} & 16.0 {\small$\pm$ 5.9\par} & 5.0 {\small$\pm$ \ \ 2.7\par} \\
LSTM-1 & 28.6 {\small$\pm$ 2.2\par} & 26.0 {\small$\pm$ 1.9\par} & 40.1 {\small$\pm$ 2.0\par} & 34.8 {\small$\pm$ 2.1\par} & 15.1 {\small$\pm$ 4.1\par} & 8.0 {\small$\pm$ \ \ 2.1\par} \\
Transformer & 28.4 {\small$\pm$ 2.5\par} & 22.1 {\small$\pm$ 2.1\par} & 48.6 {\small$\pm$ 1.7\par} & 33.9 {\small$\pm$ 3.2\par} & 45.1 {\small$\pm$ 8.0\par} & 16.4 {\small$\pm$ \ \ 8.4\par}
\end{tabular}
\end{table}

\section{Zero-Shot Learning Datasets}
\label{sec:zeroshot}
\begin{figure*}[!ht]
\centering
\subfloat[
aPY
]{
\includegraphics[width=0.48\textwidth]{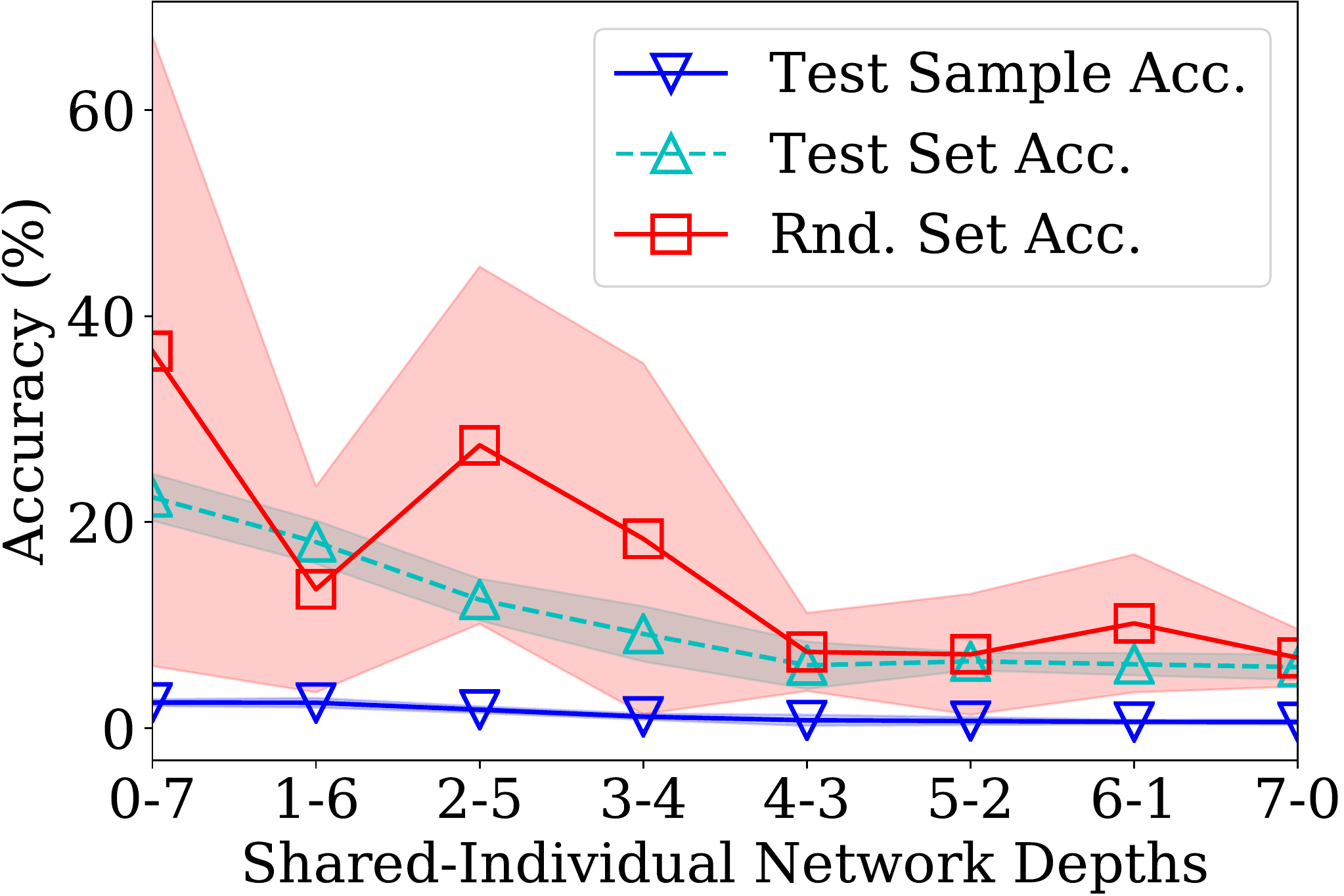}
}
\subfloat[
AwA2
]{
\includegraphics[width=0.48\textwidth]{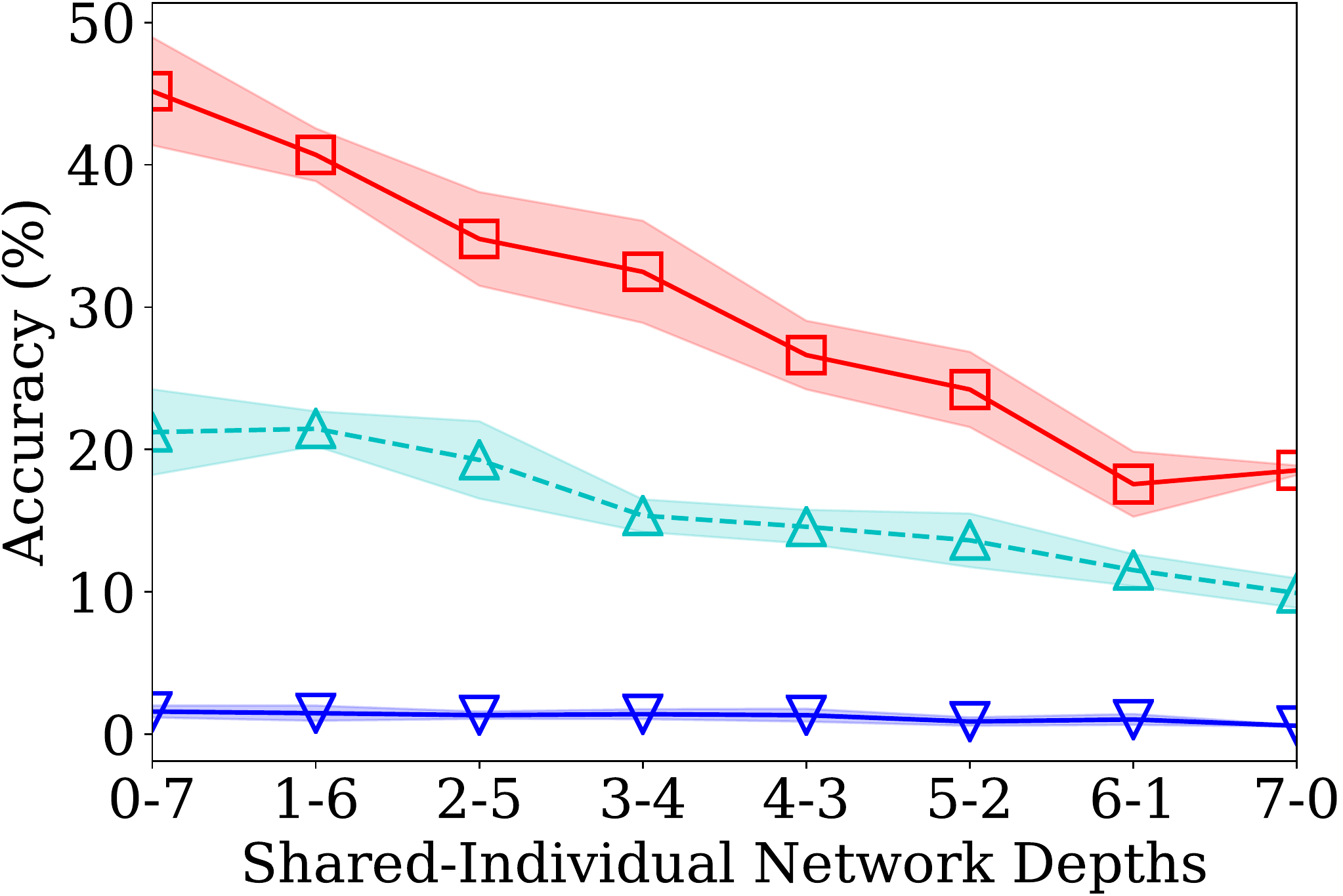}
}
\\
\subfloat[
CUB
]{
\includegraphics[width=0.48\textwidth]{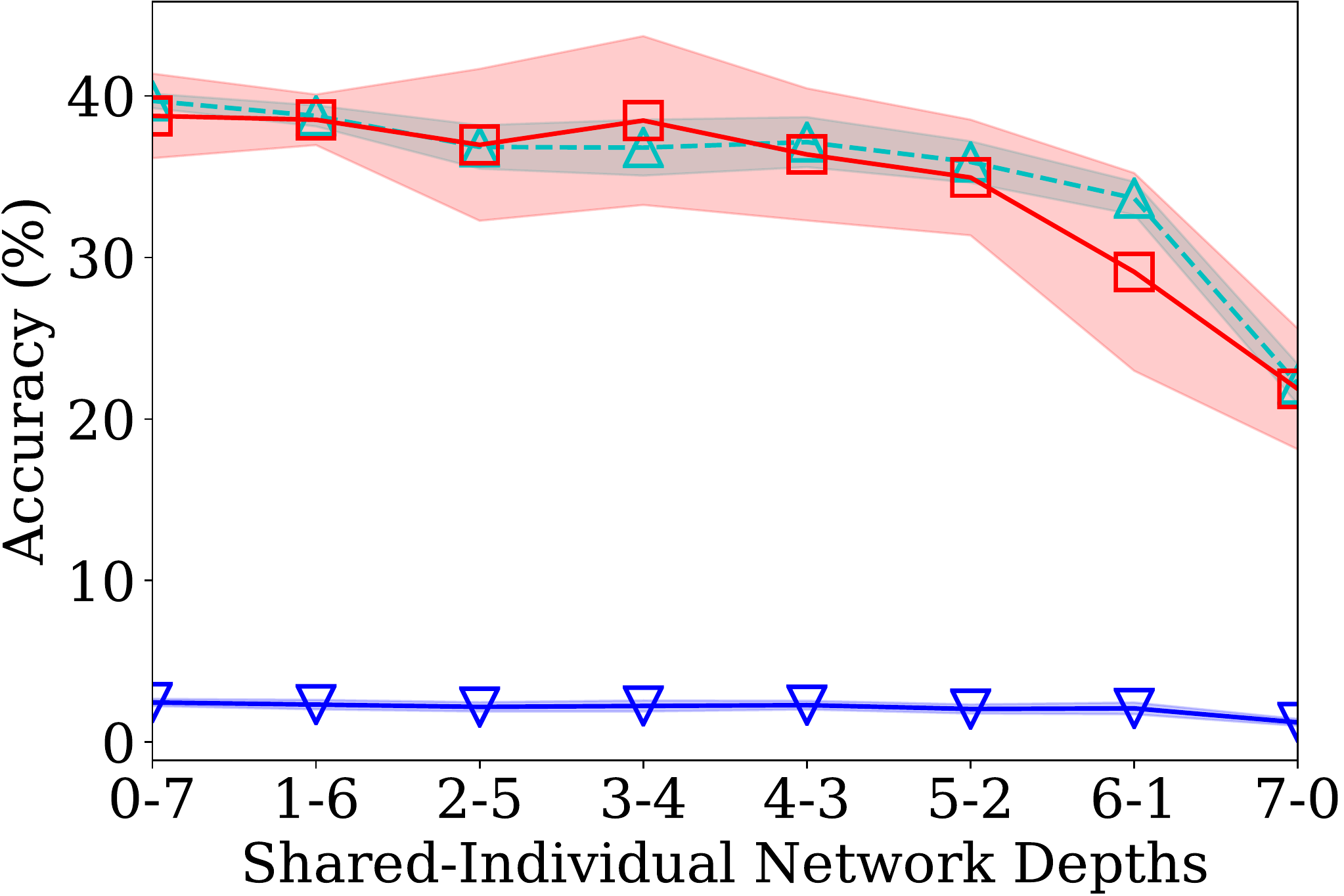}
}
\subfloat[
SUN
]{
\includegraphics[width=0.48\textwidth]{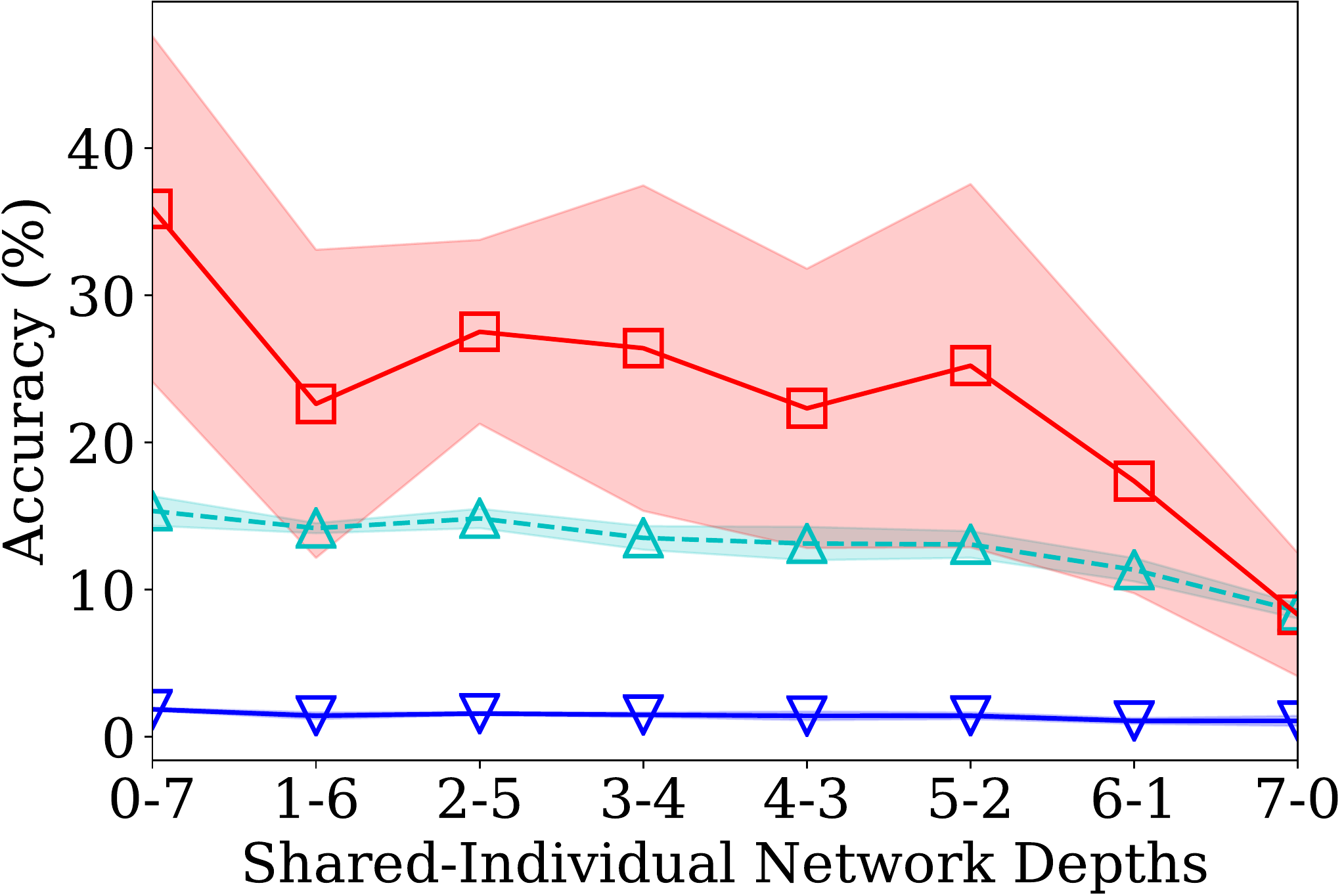}
}
\caption{
Experimental results for Zero-Shot Learning with CNN models.
The results are similar to those in the experiment section.
}
\label{fig:zeroshot_learning_results}
\end{figure*}

We look at the results for Zero-shot learning.
We use aPY~\cite{farhadi2009describing}, AwA2~\cite{xian2019zero}, CUB~\cite{wah2011the}, and SUN~\cite{patterson2012sun} datasets.
We use the pre-extracted input features for aPY and AwA and image input for CUB and SUN.
We use the most balanced attributes in all the data with a relative ratio of 0.5, three for each output.
So there are eight output nodes for each of the two outputs.
For aPY, samples may share the same image, so we construct training data from the original training and test data from the original test data.
For other datasets, we separate all the data.
We use an eight-layer CNN model, the same as in the experiment section.
The batch size is 512 for aPY and AwA and 256 for CUB and SUN.
Other settings are the same as that in the experiment section.
The result is shown in Figure~\ref{fig:zeroshot_learning_results} and Table~\ref{tab:equal_difficulty_results}.
Similar to the previous experiments, the depths of shared and individual networks influence systematic generalization.

\begin{table}[!ht]
\centering
\caption{Accuracy (mean $\pm$ std \%).
Experimental results for Zero-Shot Learning.}
\label{tab:zeroshot_learning_results}
\begin{tabular}{lrr|rr|rr}
& \multicolumn{2}{c|}{Test Sample Accuracy}
& \multicolumn{2}{c|}{Test Set Accuracy}
& \multicolumn{2}{c}{Random Set Accuracy} \\
& \multicolumn{1}{c}{Individual} & \multicolumn{1}{c|}{Shared} 
& \multicolumn{1}{c}{Individual} & \multicolumn{1}{c|}{Shared}
& \multicolumn{1}{c}{Individual} & \multicolumn{1}{c}{Shared} \\
\hline
aPY & 2.5 {\small$\pm$ 0.3\par} & 0.6 {\small$\pm$ 0.2\par} & 22.4 {\small$\pm$ 2.3\par} & 5.9 {\small$\pm$ 1.2\par} & 36.6 {\small$\pm$ 30.6\par} & 6.8 {\small$\pm$ 2.8\par} \\
AwA2 & 1.6 {\small$\pm$ 0.4\par} & 0.6 {\small$\pm$ 0.1\par} & 21.2 {\small$\pm$ 3.0\par} & 9.9 {\small$\pm$ 1.0\par} & 45.2 {\small$\pm$ \ \ 3.8\par} & 18.5 {\small$\pm$ 0.3\par} \\
CUB & 2.5 {\small$\pm$ 0.2\par} & 1.2 {\small$\pm$ 0.2\par} & 39.7 {\small$\pm$ 0.4\par} & 22.2 {\small$\pm$ 1.2\par} & 38.8 {\small$\pm$ \ \ 2.6\par} & 21.9 {\small$\pm$ 3.7\par} \\
SUN & 1.9 {\small$\pm$ 0.1\par} & 1.1 {\small$\pm$ 0.3\par} & 15.3 {\small$\pm$ 1.0\par} & 8.5 {\small$\pm$ 0.5\par} & 35.9 {\small$\pm$ 11.7\par} & 8.3 {\small$\pm$ 4.2\par}
\end{tabular}
\end{table}

\section{More Discussions}
\label{sec:more_discussion}

\subsection{Training process}
\begin{figure*}[!ht]
\centering
\subfloat[
DNN o.o.d. partition number
]{
\includegraphics[width=0.48\textwidth]{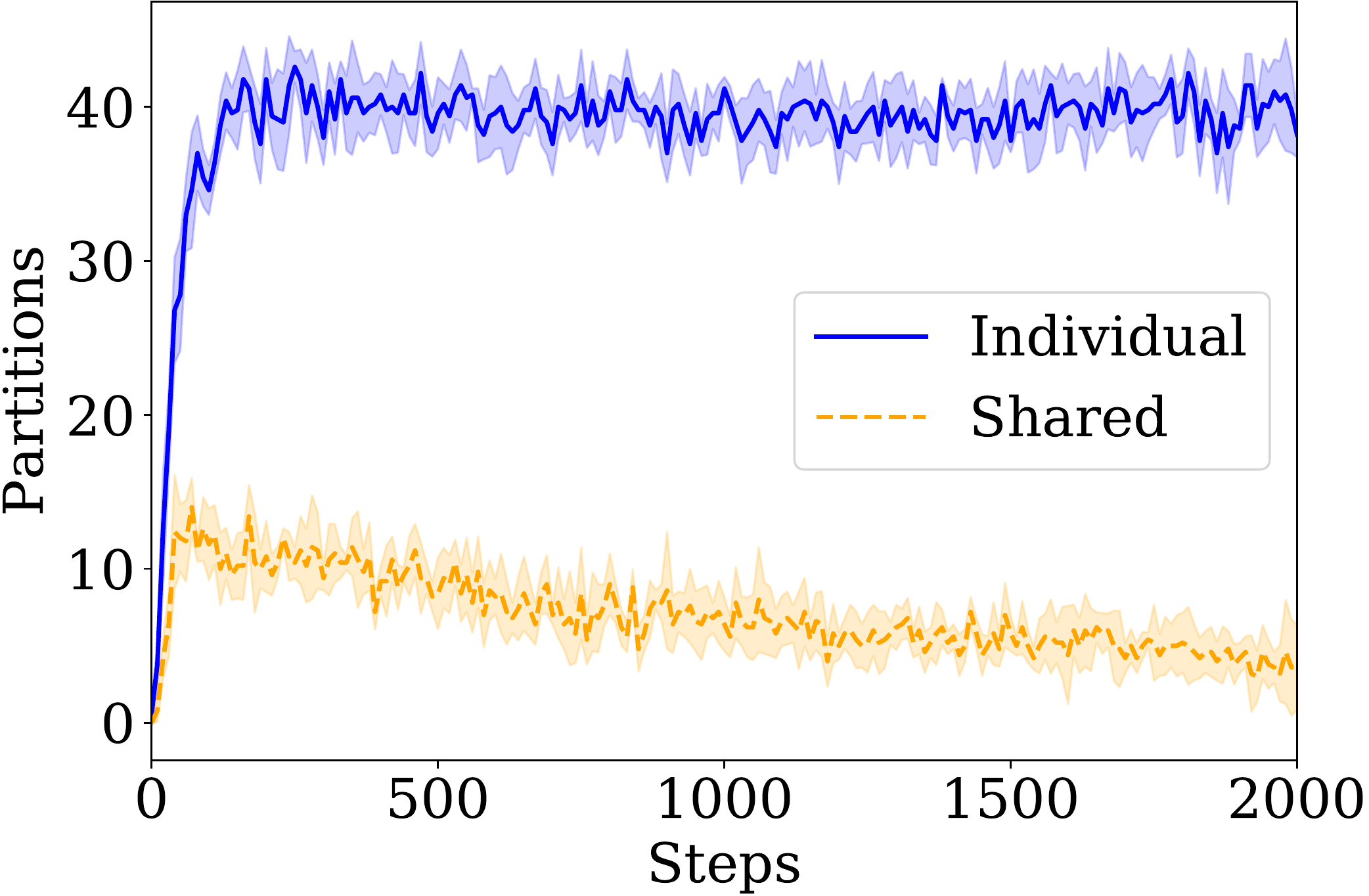}
}
\subfloat[
DNN o.o.d. test sample ratio
]{
\includegraphics[width=0.48\textwidth]{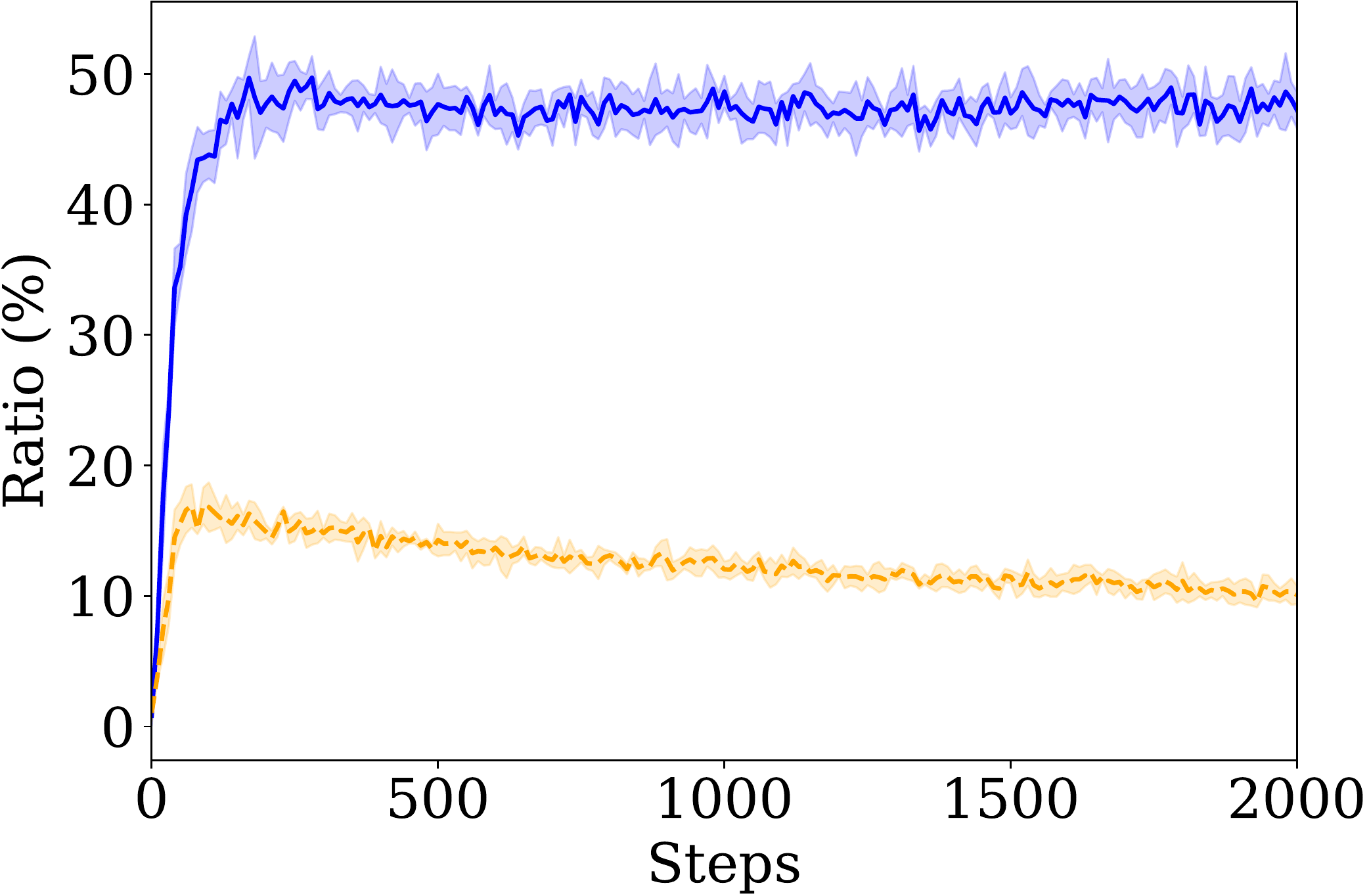}
}
\\
\subfloat[
CNN o.o.d. partition number
]{
\includegraphics[width=0.48\textwidth]{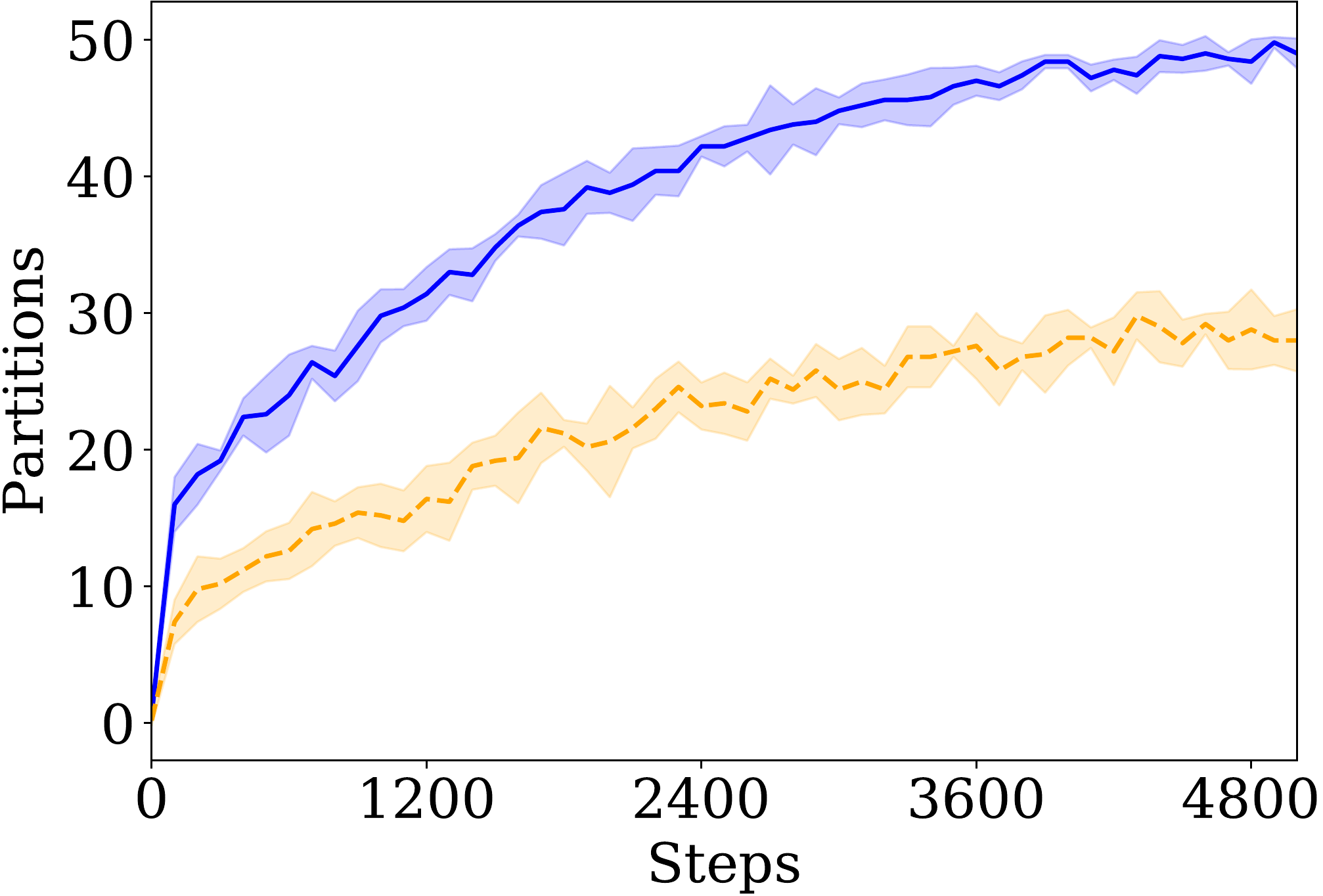}
}
\subfloat[
CNN o.o.d. test sample ratio
]{
\includegraphics[width=0.48\textwidth]{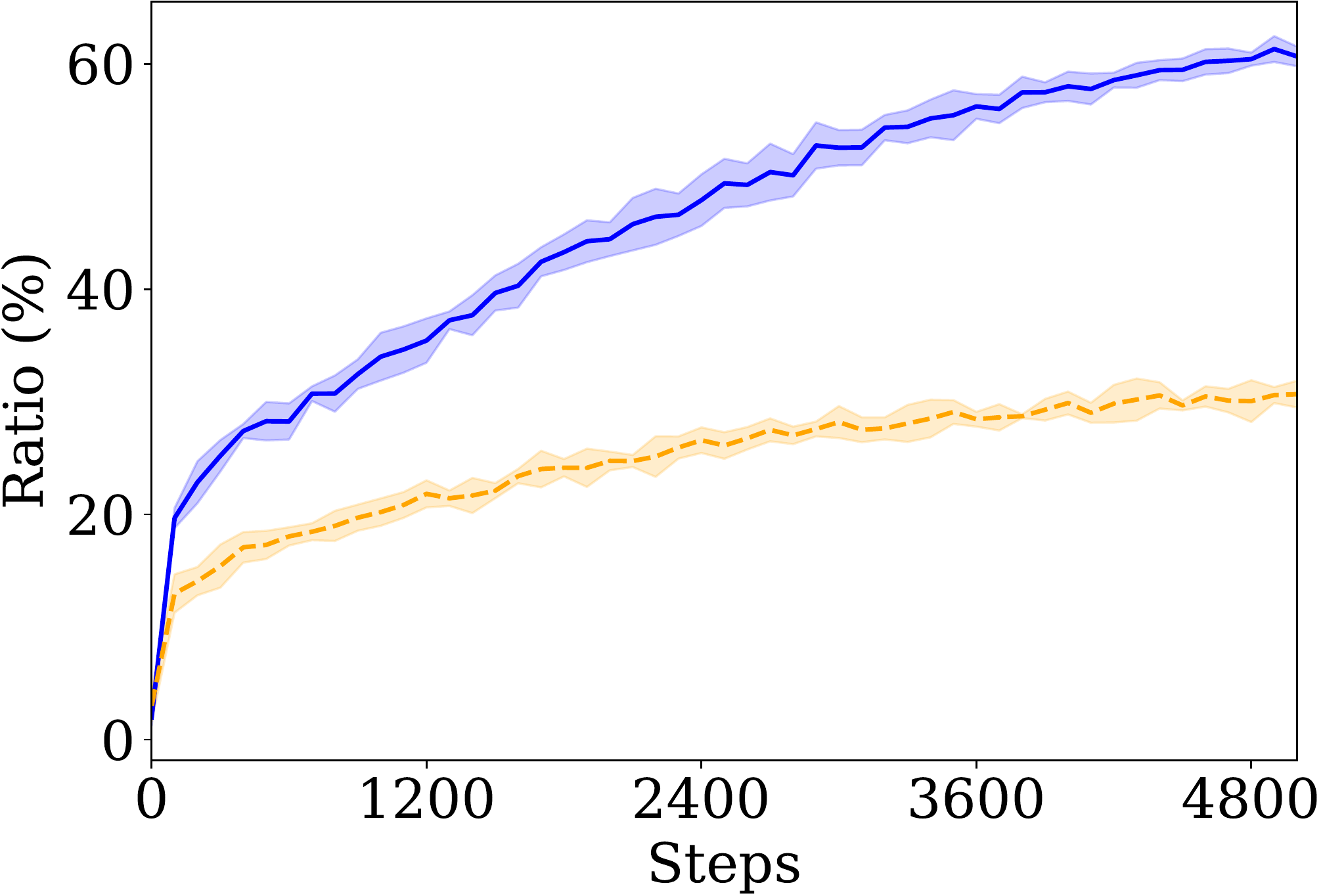}
}
\caption{
Training process.
The o.o.d. partition number and the o.o.d. test sample ratio for shared and individual networks.
}
\label{fig:training_process}
\end{figure*}

We discussed that sharing boundaries reduces the number of partitions and shrinks the area for new outputs (Proposition~\ref{prop:seen_prediction}).
We run experiments to find when this happens during training.
We sample 10,000 inputs from test data, and if an output combination has at least 50 samples, we regard it as a new output (o.o.d.) partition.
We plot the number of o.o.d. partitions and the test sample ratio in the o.o.d. partitions for shared and individual networks in Figure~\ref{fig:training_process}.
The experiment settings follow DNN and CNN settings in the experiment section.
The results show that the differences start to happen in early training.

\subsection{New classes}
In the experiment section, a new output is the recombination of seen factors.
It is also available to design experiments for new classes without factor combinations.
However, it might be less attractive because tasks usually do not expect a model to predict a new class unrelated to seen classes.
Also, the model design needs to meet some requirements.

We look at ten-class classification problems.
We use the first five classes in the training and the remaining five classes in the test.
For a fully connected network, we use the Fashion dataset.
For a convolutional neural network, we use the CIFAR-10 dataset.
The networks do not have biases.
There are ten pieces of individual networks for a model.
We use binary cross-entropy for each output node and sum them as the loss.
Other architecture settings are the same as those in the experiment section.

The results are in Table~\ref{tab:single}.
The individual network has non-zero random set accuracy, which means the input space has elements mapped to new outputs.
The shared network does not have such inputs.
It indicates that the sharing of functions discourages predicting the new output.

\begin{table}[!ht]
\centering
\caption{Accuracy (mean $\pm$ std \%).
Experimental results for single output.}
\label{tab:single}
\begin{tabular}{lrr|rr|rr}
& \multicolumn{2}{c|}{Test Sample Accuracy}
& \multicolumn{2}{c|}{Test Set Accuracy}
& \multicolumn{2}{c}{Random Set Accuracy} \\
& \multicolumn{1}{c}{Individual} & \multicolumn{1}{c|}{Shared} 
& \multicolumn{1}{c}{Individual} & \multicolumn{1}{c|}{Shared}
& \multicolumn{1}{c}{Individual} & \multicolumn{1}{c}{Shared} \\
\hline
DNN & 0.0 {\small$\pm$ 0.0\par} & 0.0 {\small$\pm$ 0.0\par} & 0.0 {\small$\pm$ 0.0\par} & 0.0 {\small$\pm$ 0.0\par} & 3.8 {\small$\pm$ 1.7\par} & 0.0 {\small$\pm$ 0.0\par} \\
CNN & 0.0 {\small$\pm$ 0.0\par} & 0.0 {\small$\pm$ 0.0\par} & 0.0 {\small$\pm$ 0.0\par} & 0.0 {\small$\pm$ 0.0\par} & 0.0 {\small$\pm$ 0.1\par} & 0.0 {\small$\pm$ 0.0\par}
\end{tabular}
\end{table}

\subsection{Illustrative examples}
\begin{figure*}[!ht]
\centering
\subfloat[
Preferred model.
]{
\includegraphics[width=0.23\textwidth]{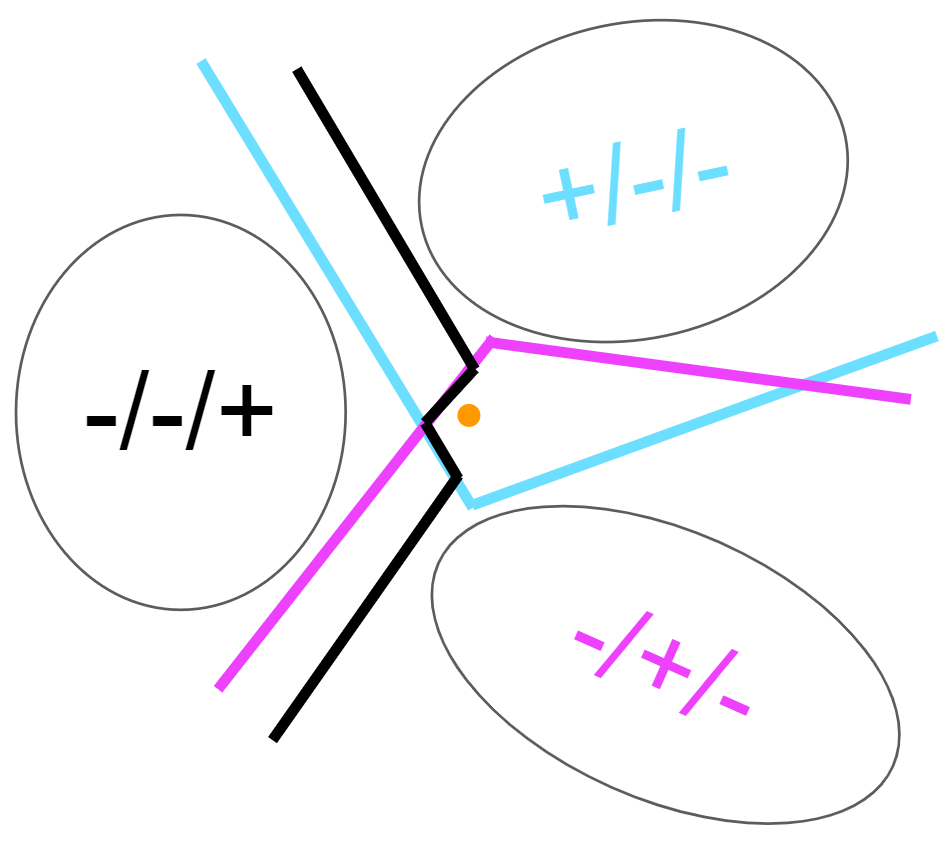}
\label{fig:conceptual_three_preferred}
}
\subfloat[
Not preferred model.
]{
\includegraphics[width=0.23\textwidth]{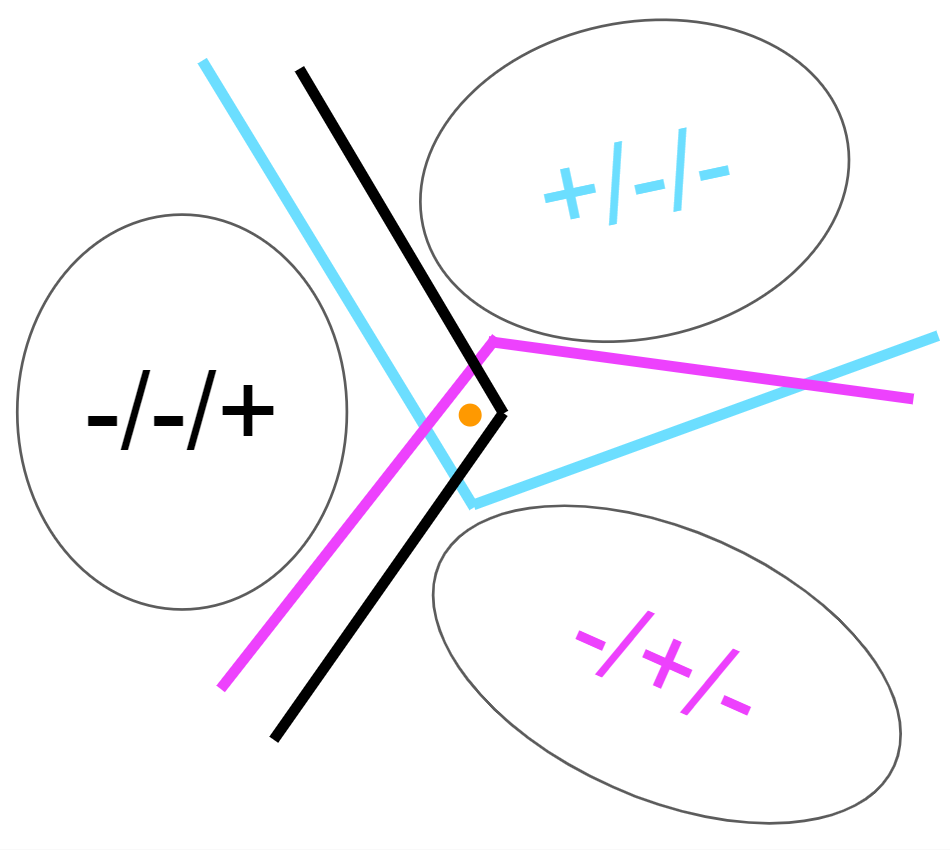}
\label{fig:conceptual_three_unpreferred}
}
\caption{
Intuitive examples for three factors.
The input space contains three manifolds: +/-/-, -/+/-, and -/-/+.
The output has two factors, and the model learns three decision boundaries (cyan, pink and black).
The orange dot is a test sample with a new factor combination +/+/+.
We argue that there is a bias to prefer the first case (a) over the second case (b).
It is because the training process tries to reuse functions.
Hence, it avoids systematic generalization.
}
\label{fig:concepts_three}
\end{figure*}

We have mainly used examples for two factors.
Figure~\ref{fig:concepts_three} is an illustration of $K>2$ factors.

\subsection{Examples in other conditions}
\label{sec:more_explanatory_examples}
\begin{figure*}[!ht]
\captionsetup[subfigure]{labelformat=empty}
\centering
\subfloat[
(a) First output
]{
\includegraphics[width=0.145\textwidth]{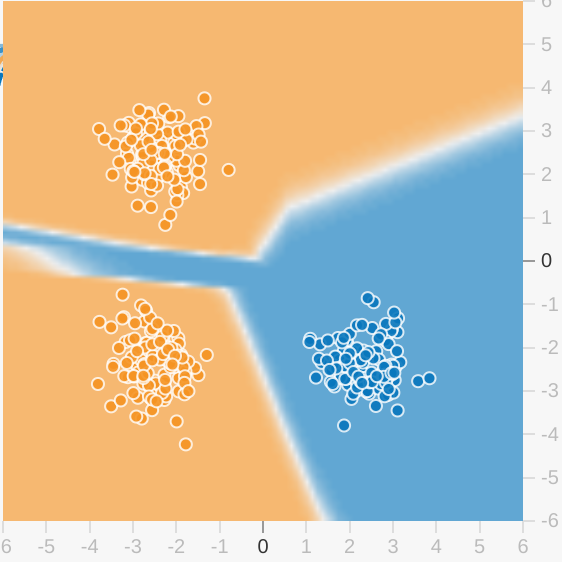}
}
\subfloat[
Second output
]{
\includegraphics[width=0.145\textwidth]{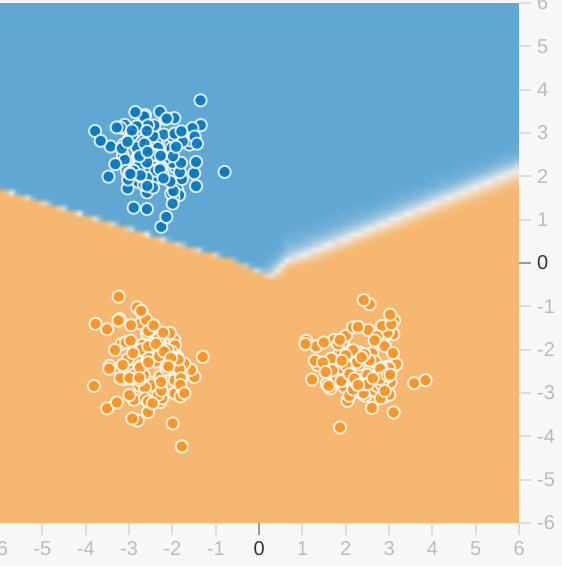}
}
\subfloat[
Result
]{
\includegraphics[width=0.145\textwidth]{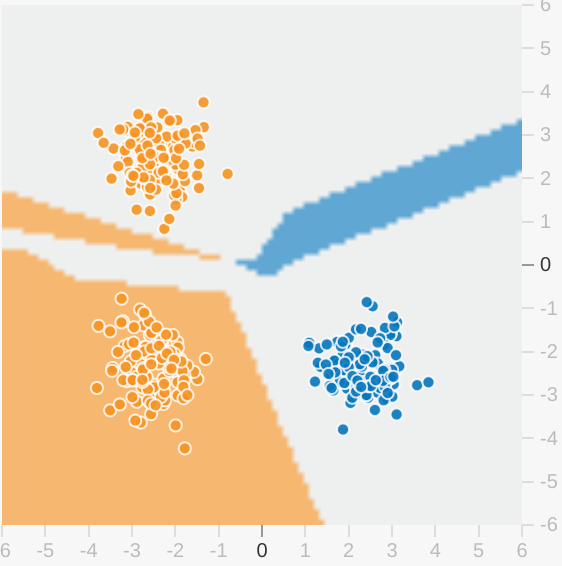}
}
\quad
\subfloat[
(b) First output
]{
\includegraphics[width=0.145\textwidth]{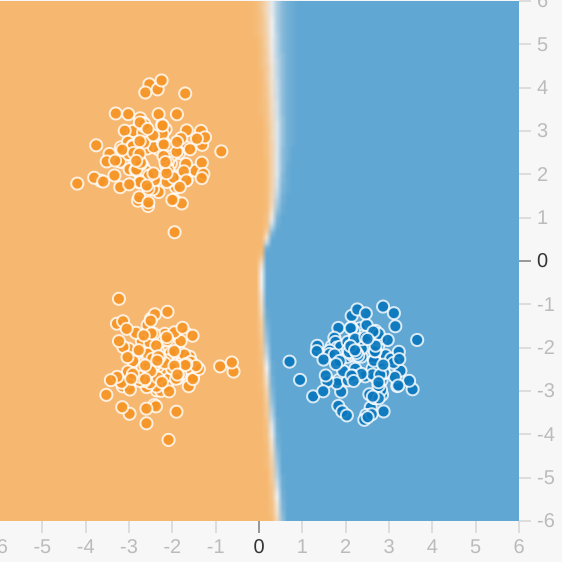}
}
\subfloat[
Second output
]{
\includegraphics[width=0.145\textwidth]{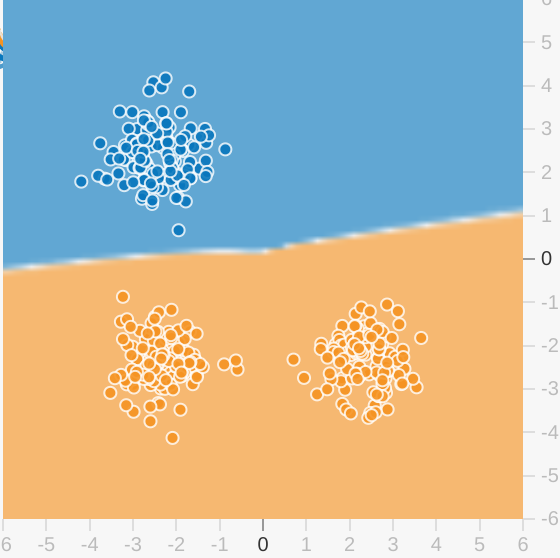}
}
\subfloat[
Result
]{
\includegraphics[width=0.145\textwidth]{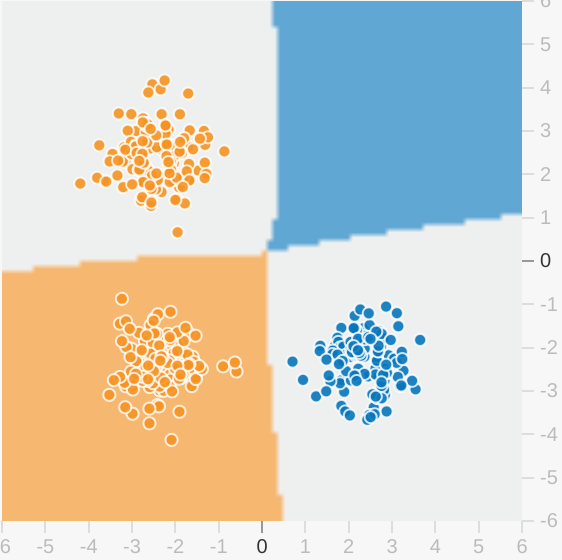}
}
\caption{
Examples in other conditions.
(a) has a shallow network.
The new combination is enabled because the last layer has different parameters.
Note that the upper and lower boundaries are parallel.
(b) uses hyperbolic tangent (tanh) activation, similar to the linear model when the values are close to zero and the distribution is a specific linear one.
}
\label{fig:counter_examples}
\end{figure*}

We focus on solving complicated non-linear problems with deep learning models.
Figure~\ref{fig:counter_examples} shows examples not in such condition.
One example (a) is a shallow network.
Another example (b) has a linear problem, and the model design matches it.

\subsection{More visualized examples}
\label{sec:more_visualization_example}
\begin{figure*}[!ht]
\captionsetup[subfigure]{labelformat=empty}
\centering
\subfloat[
(a) First output
]{
\includegraphics[width=0.15\textwidth]{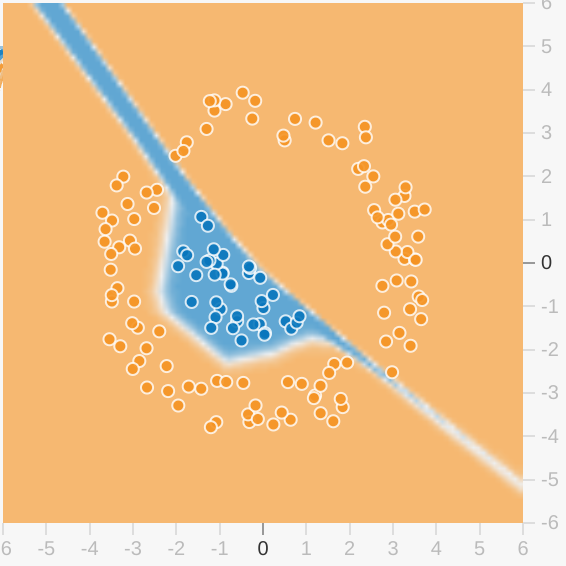}
}
\subfloat[
Second output
]{
\includegraphics[width=0.15\textwidth]{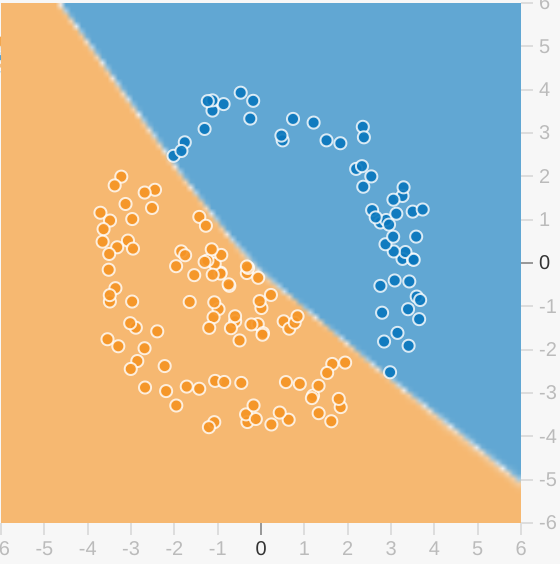}
}
\subfloat[
Result
]{
\includegraphics[width=0.15\textwidth]{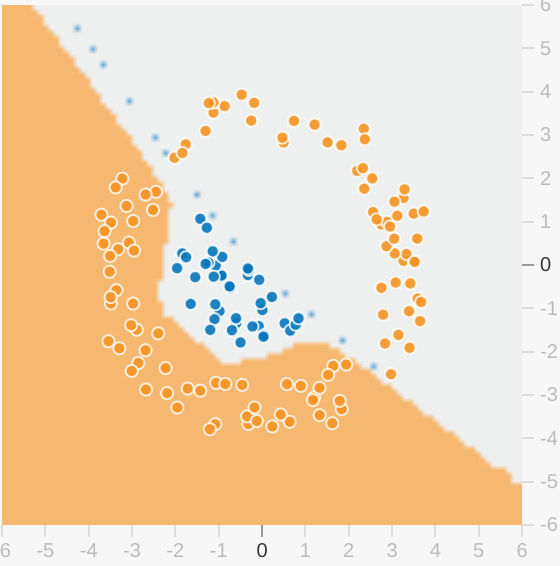}
}
\quad
\subfloat[
(b) First output
]{
\includegraphics[width=0.15\textwidth]{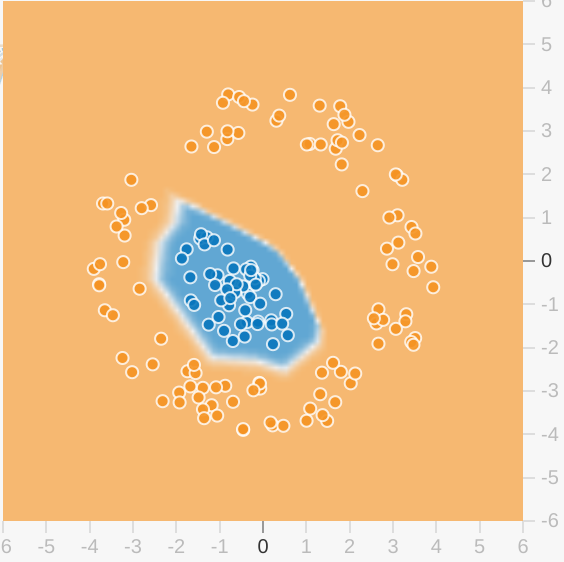}
}
\subfloat[
Second output
]{
\includegraphics[width=0.15\textwidth]{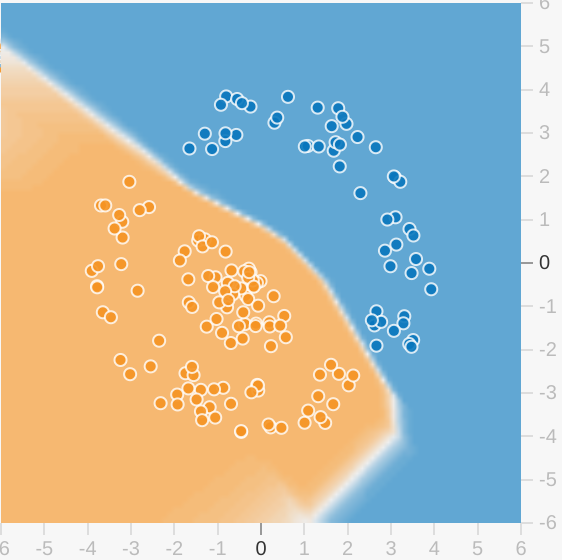}
}
\subfloat[
Result
]{
\includegraphics[width=0.145\textwidth]{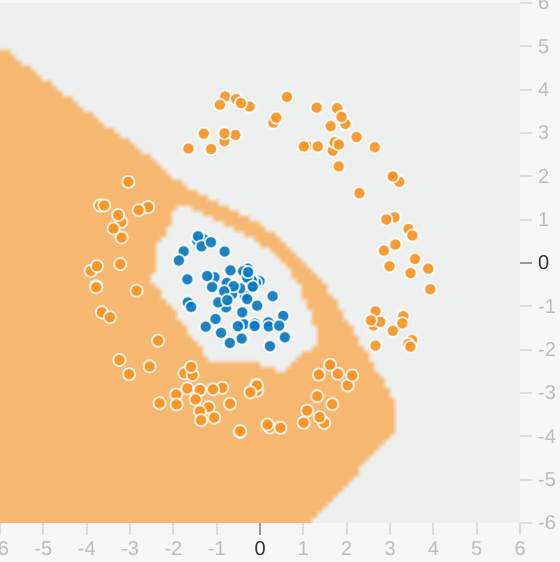}
}
\\
\subfloat[
(c) First output
]{
\includegraphics[width=0.15\textwidth]{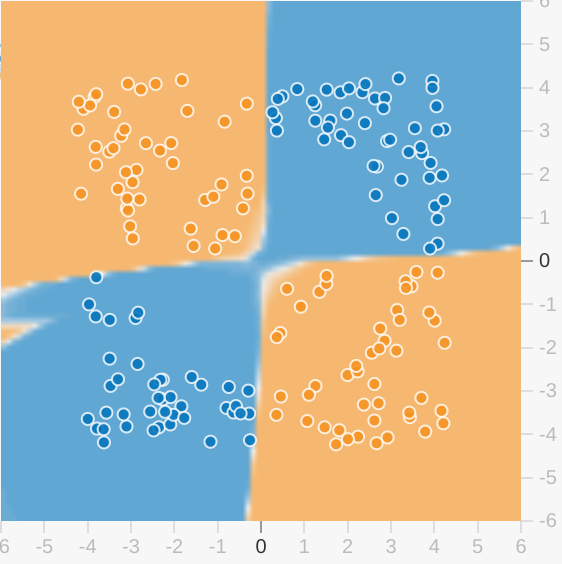}
}
\subfloat[
Second output
]{
\includegraphics[width=0.15\textwidth]{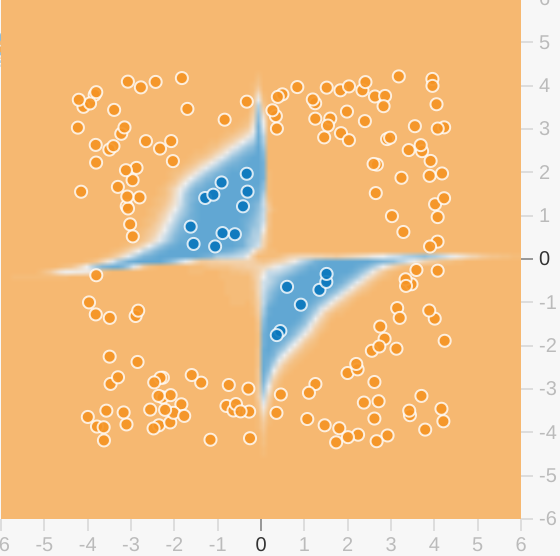}
}
\subfloat[
Result
]{
\includegraphics[width=0.145\textwidth]{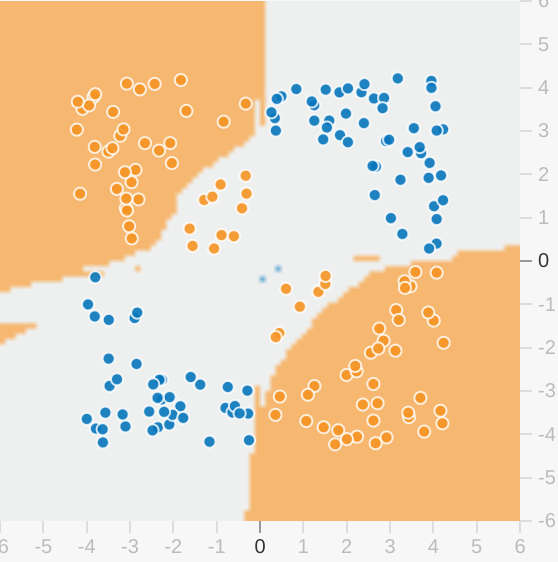}
}
\quad
\subfloat[
(d) First output
]{
\includegraphics[width=0.15\textwidth]{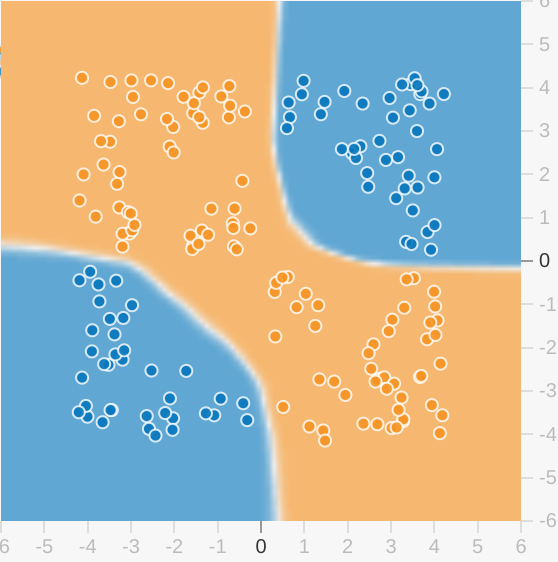}
}
\subfloat[
Second output
]{
\includegraphics[width=0.15\textwidth]{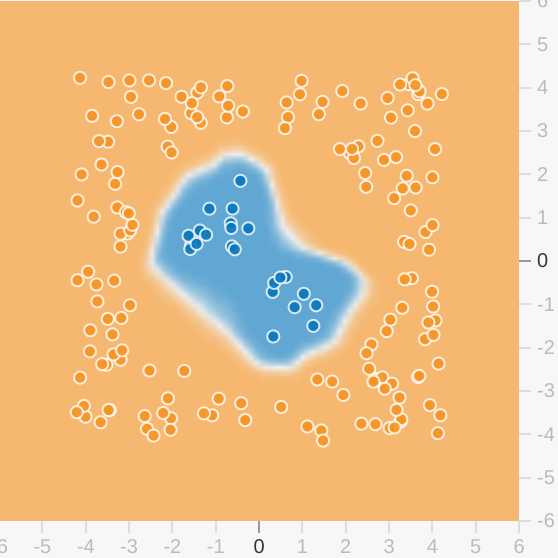}
}
\subfloat[
Result
]{
\includegraphics[width=0.145\textwidth]{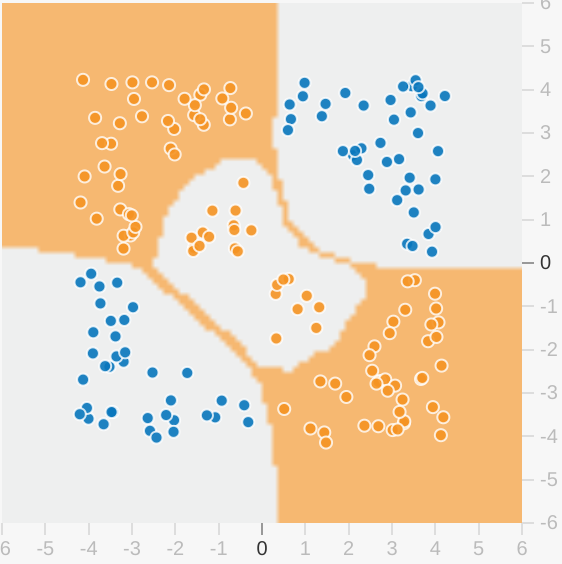}
}
\\
\subfloat[
(e) First output
]{
\includegraphics[width=0.15\textwidth]{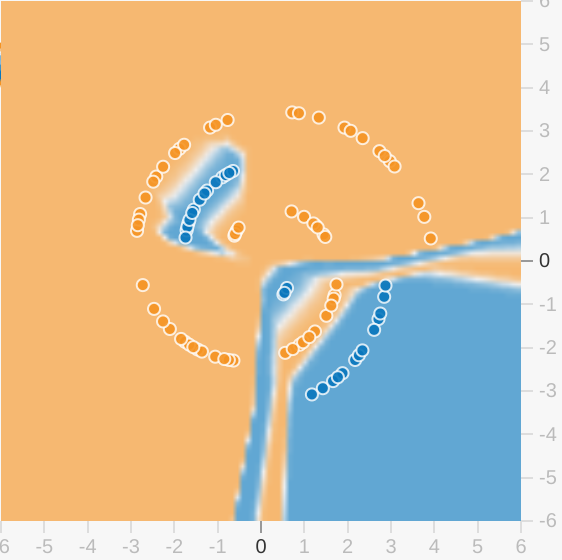}
}
\subfloat[
Second output
]{
\includegraphics[width=0.15\textwidth]{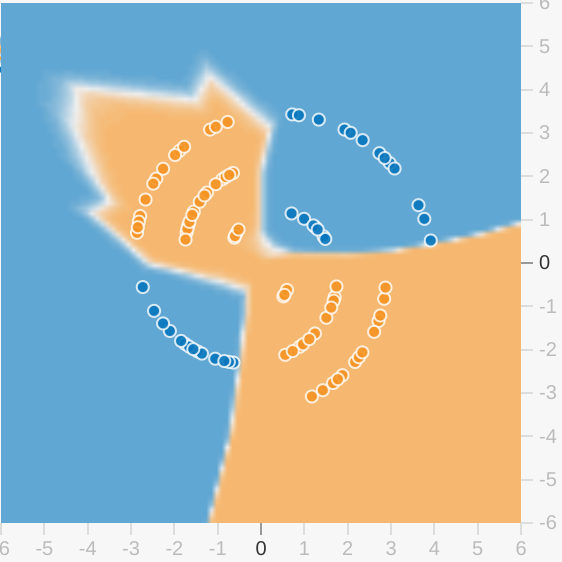}
}
\subfloat[
Result
]{
\includegraphics[width=0.145\textwidth]{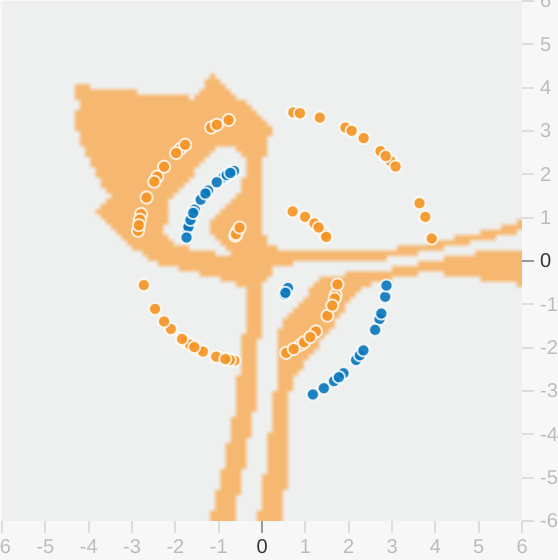}
}
\quad
\subfloat[
(f) First output
]{
\includegraphics[width=0.15\textwidth]{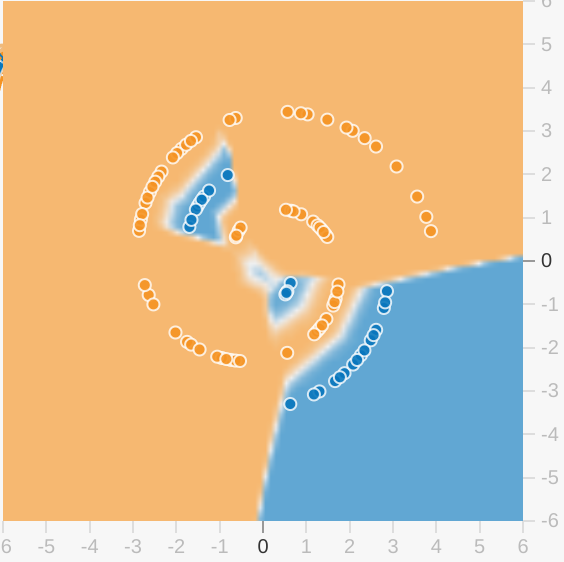}
}
\subfloat[
Second output
]{
\includegraphics[width=0.15\textwidth]{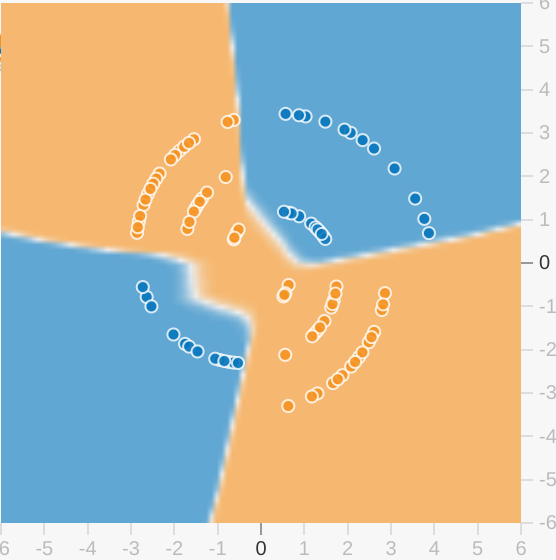}
}
\subfloat[
Result
]{
\includegraphics[width=0.145\textwidth]{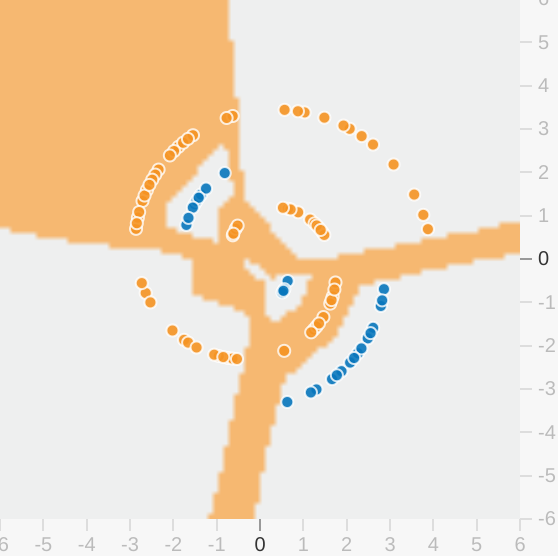}
}
\\
\subfloat[
(g) First output
]{
\includegraphics[width=0.15\textwidth]{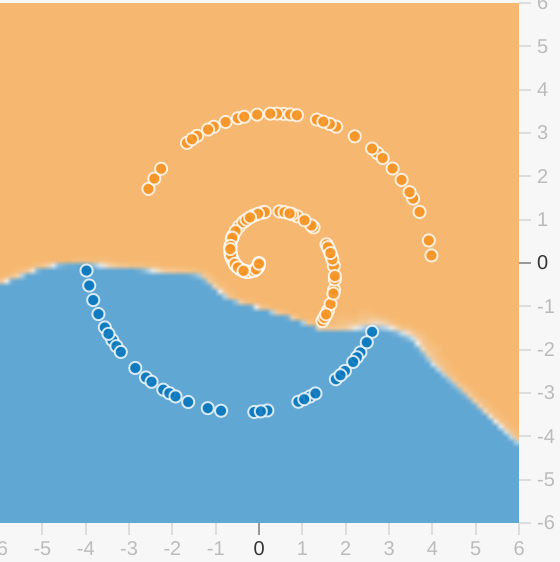}
}
\subfloat[
Second output
]{
\includegraphics[width=0.15\textwidth]{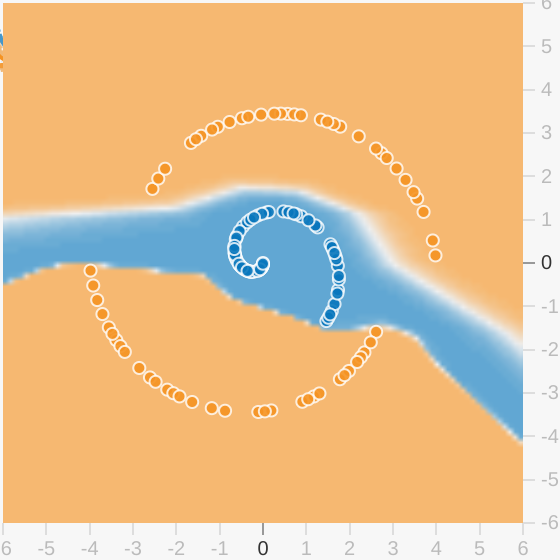}
}
\subfloat[
Result
]{
\includegraphics[width=0.145\textwidth]{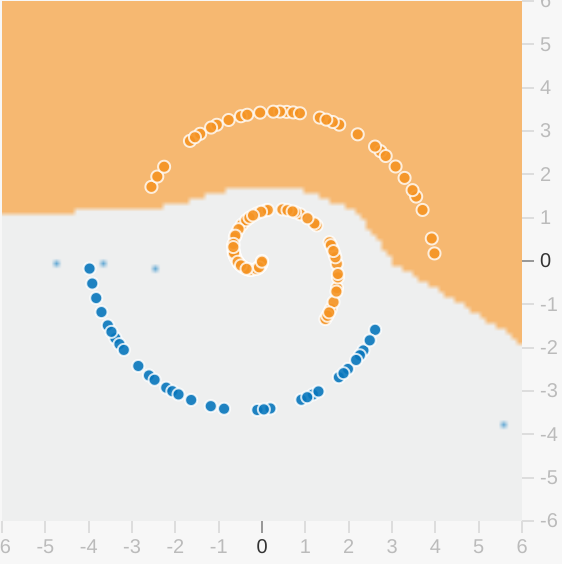}
}
\quad
\subfloat[
(h) First output
]{
\includegraphics[width=0.15\textwidth]{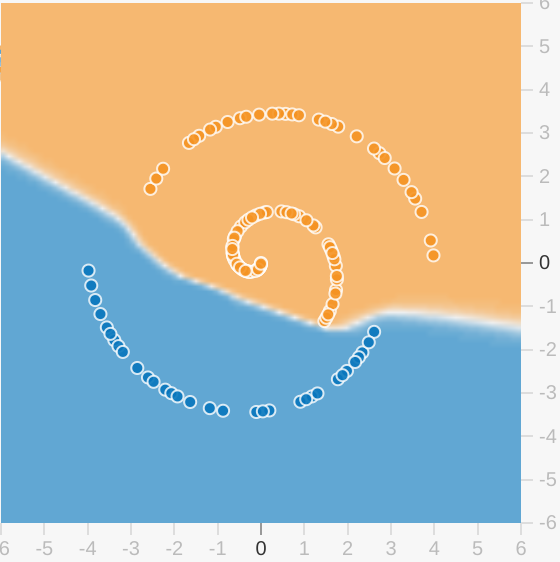}
}
\subfloat[
Second output
]{
\includegraphics[width=0.15\textwidth]{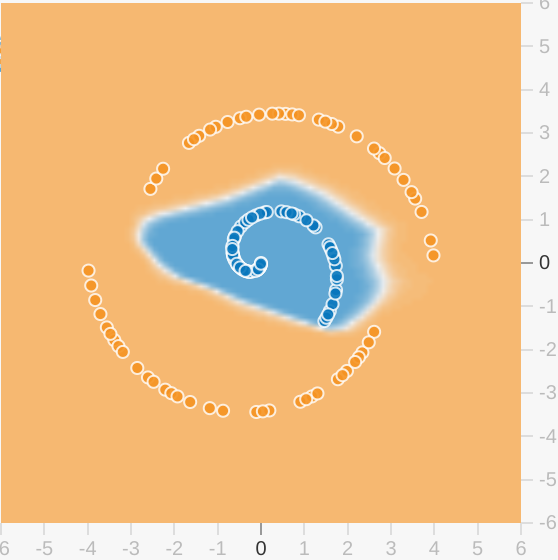}
}
\subfloat[
Result
]{
\includegraphics[width=0.145\textwidth]{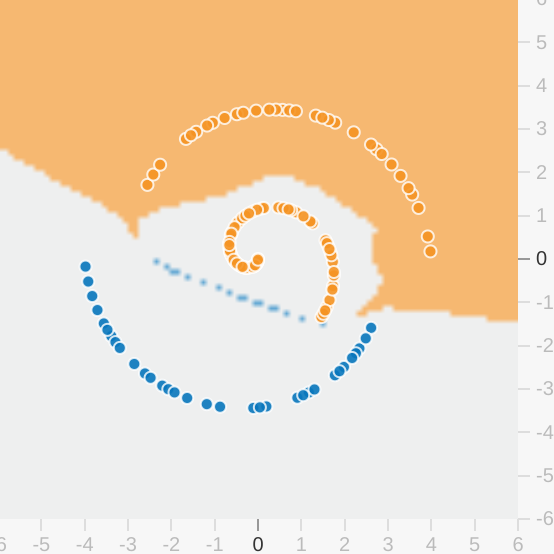}
}
\caption{
Additional visualized examples.
The settings are the same as those in Figure~\ref{fig:explanatory_examples}.
Blue overlapping areas, if any, only exist at boundaries.
}
\label{fig:additional_explanatory_examples}
\end{figure*}

We look at the visualized example.
We use a more complicated example than the one in the previous section.
The first output is spiral, and the second output is XOR.
The result shows that all three accuracies tend to drop as the shared network depth increases.
The result is in Figure~\ref{fig:additional_explanatory_examples}.

\subsection{Equally difficult factors}
\begin{figure*}[!ht]
\centering
\subfloat[
DNN
]{
\includegraphics[width=0.48\textwidth]{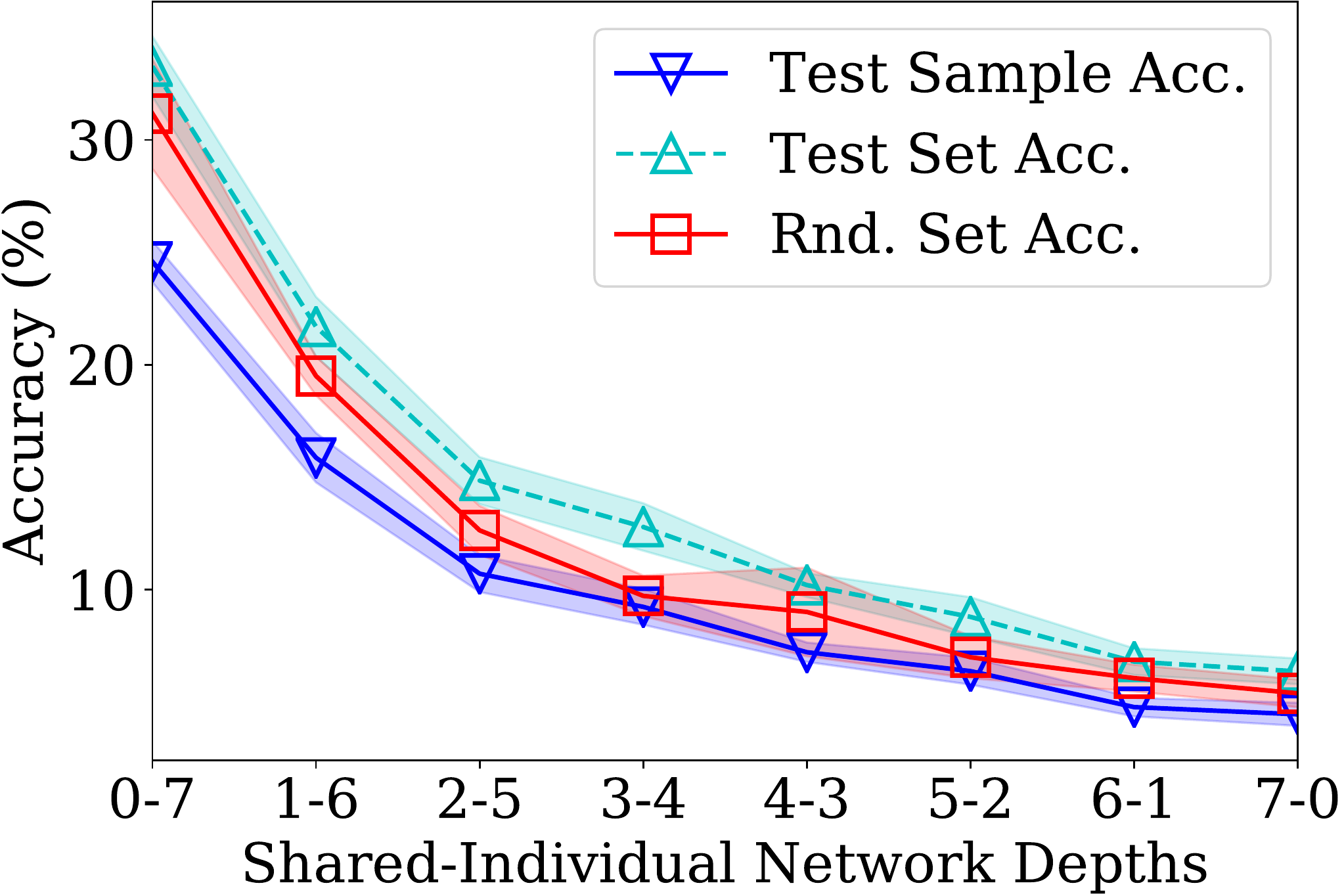}
}
\subfloat[
CNN
]{
\includegraphics[width=0.48\textwidth]{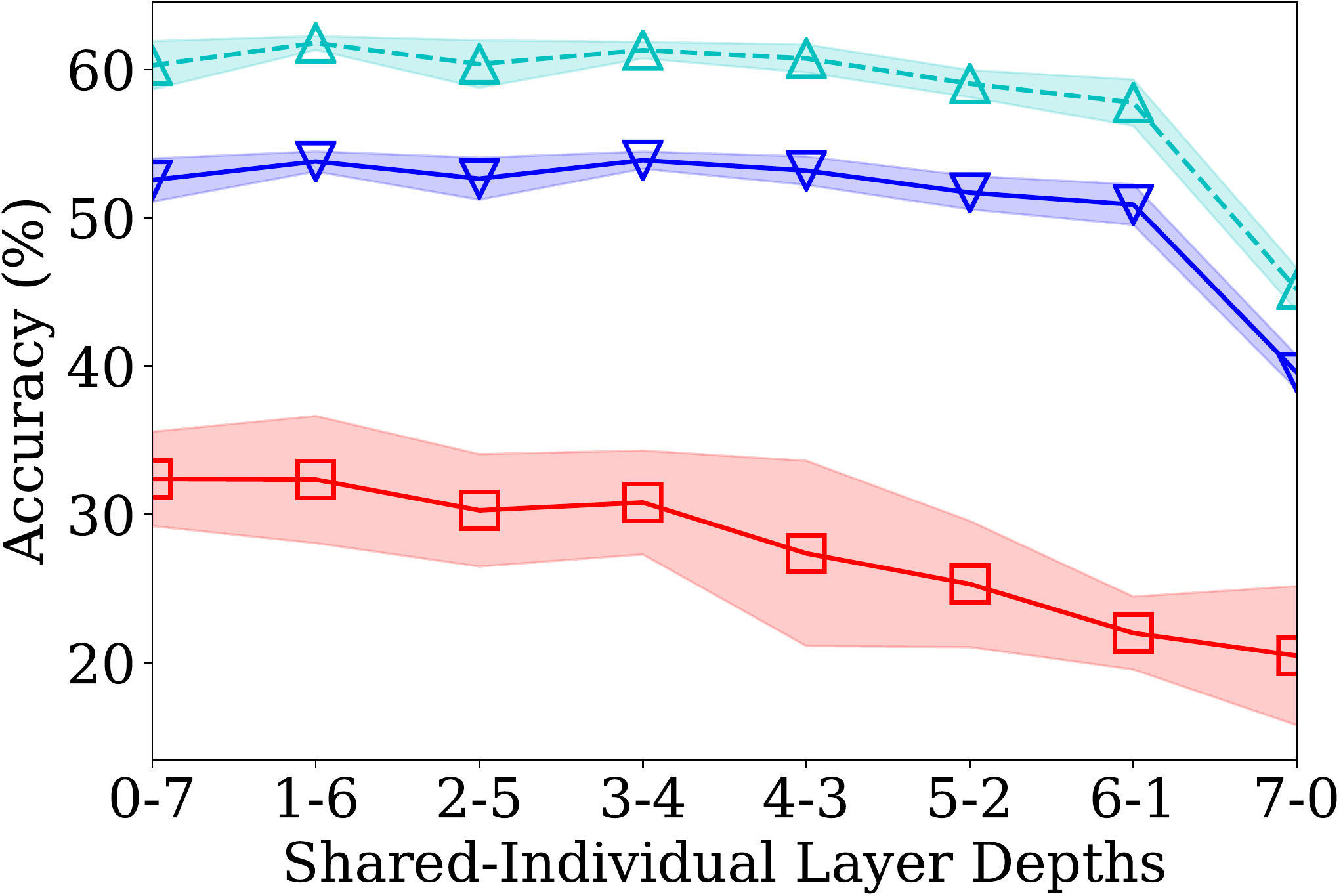}
}
\caption{
Experimental results for equally difficult factors.
The results are similar to those in the experiment section.
}
\label{fig:equal_difficulty_results}
\end{figure*}

We look at the results when two inputs are equally hard to learn.
We use two CIFAR-10 datasets for both the first and the second datasets.
Since it is too hard to learn added samples, we merge the samples by concatenating them by channel.
It means the input channel is six.
The result is shown in Figure~\ref{fig:equal_difficulty_results} and Table~\ref{tab:equal_difficulty_results}.
Similar to the previous experiments, when the difficulties are equal, the depth of the shared network influences systematic generalization.

\begin{table}[!ht]
\centering
\caption{Accuracy (mean $\pm$ std \%).
Experimental results for equally difficult factors.}
\label{tab:equal_difficulty_results}
\begin{tabular}{lrr|rr|rr}
& \multicolumn{2}{c|}{Test Sample Accuracy}
& \multicolumn{2}{c|}{Test Set Accuracy}
& \multicolumn{2}{c}{Random Set Accuracy} \\
& \multicolumn{1}{c}{Individual} & \multicolumn{1}{c|}{Shared} 
& \multicolumn{1}{c}{Individual} & \multicolumn{1}{c|}{Shared}
& \multicolumn{1}{c}{Individual} & \multicolumn{1}{c}{Shared} \\
\hline
DNN & 24.6 {\small$\pm$ 0.9\par} & 4.4 {\small$\pm$ 0.5\par} & 33.3 {\small$\pm$ 1.3\par} & 6.4 {\small$\pm$ 0.6\par} & 31.2 {\small$\pm$ 2.5\par} & 5.4 {\small$\pm$ 0.6\par} \\
CNN & 52.5 {\small$\pm$ 1.5\par} & 39.6 {\small$\pm$ 1.2\par} & 60.3 {\small$\pm$ 1.6\par} & 45.2 {\small$\pm$ 1.5\par} & 32.4 {\small$\pm$ 3.2\par} & 20.5 {\small$\pm$ 4.7\par}
\end{tabular}
\end{table}

\subsection{Label combinations}
\begin{figure*}[!ht]
\centering
\subfloat[
Tile 
]{
\includegraphics[width=0.48\textwidth]{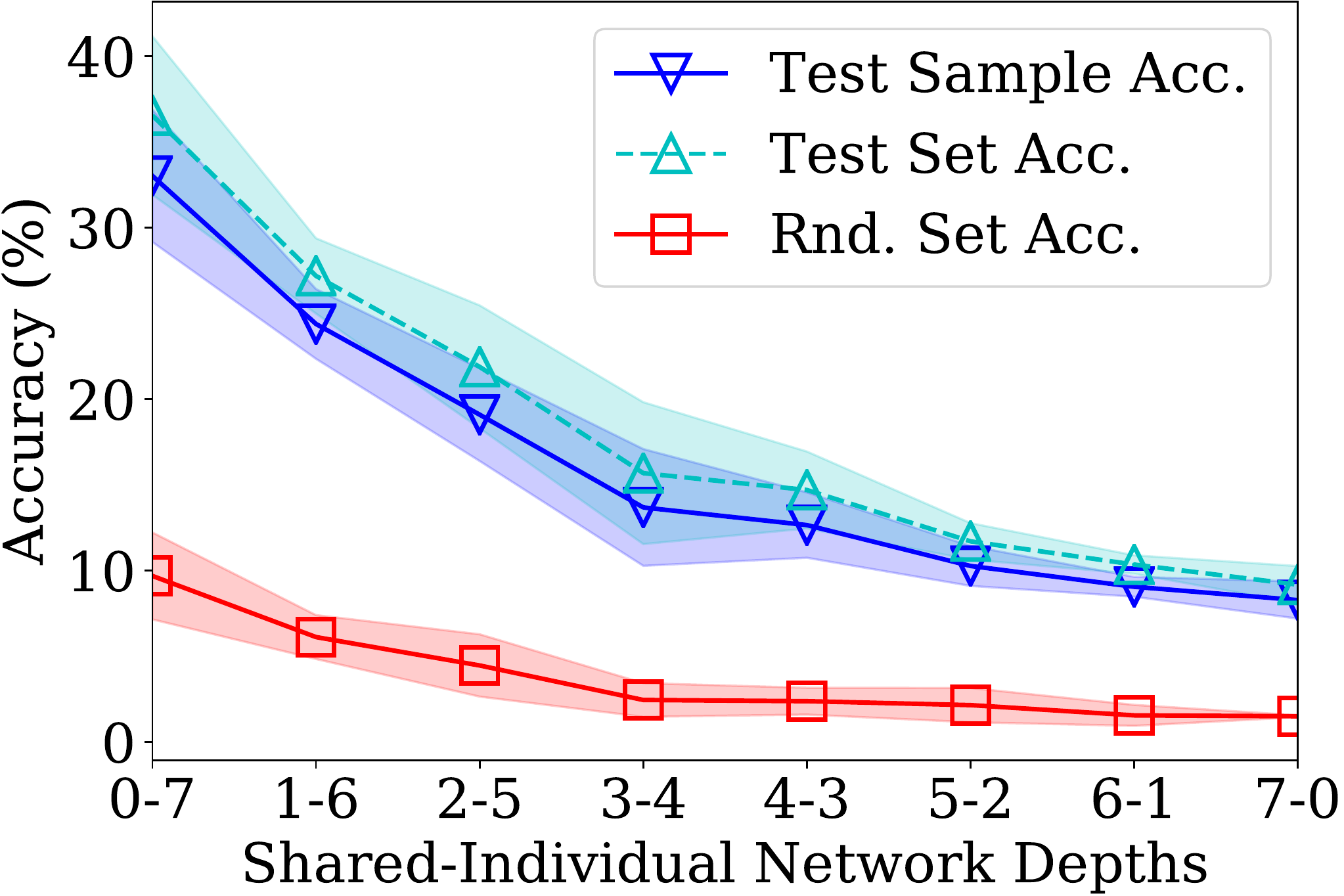}
\label{fig:cnn_tile}
}
\subfloat[
One-shot
]{
\includegraphics[width=0.48\textwidth]{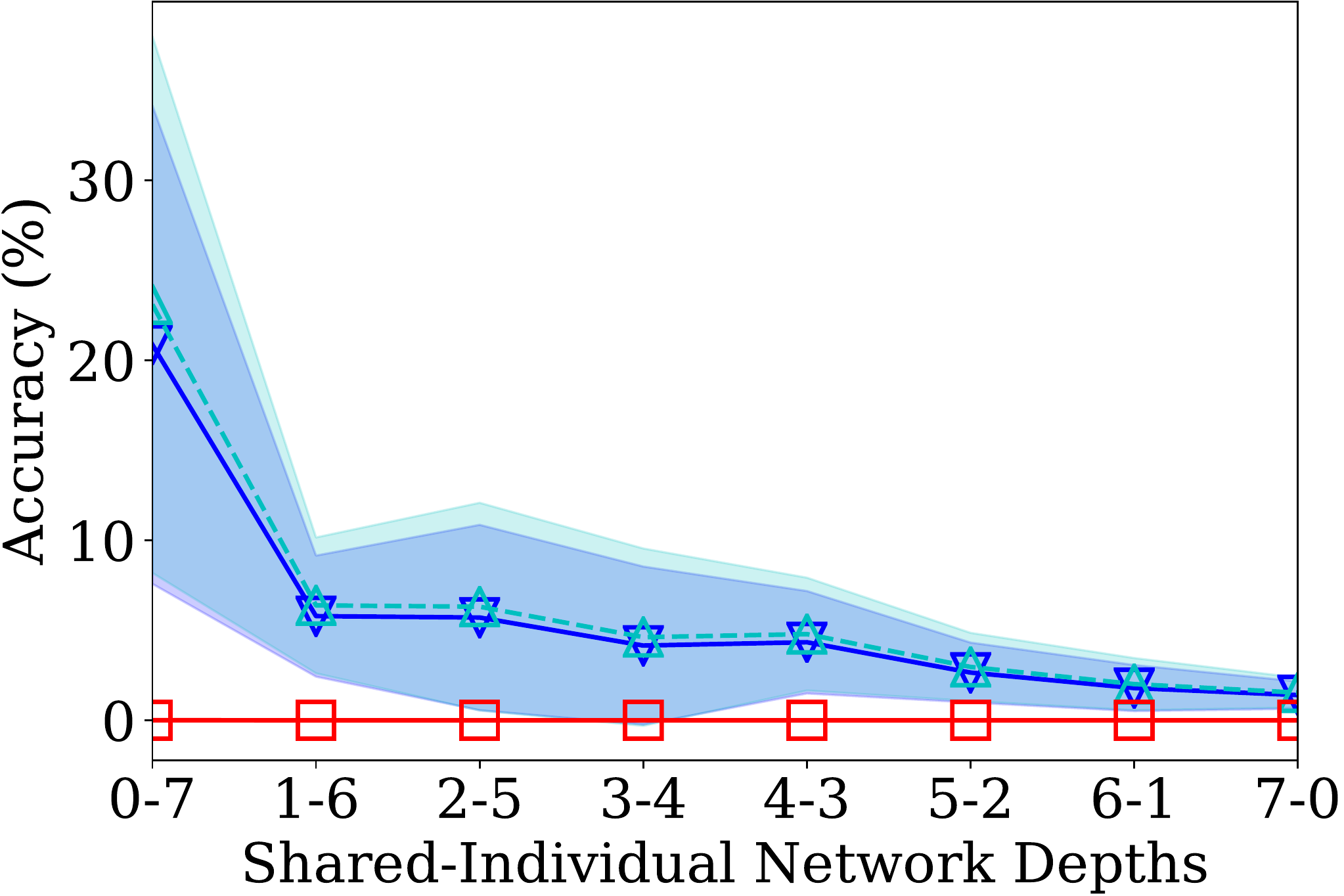}
\label{fig:cnn_oneshot}
}
\caption{
Experimental results for other label combinations.
The results are similar to those in the experiment section.
}
\label{fig:label_combination_results}
\end{figure*}

We also test other training label distribution types.
We design tile and one-label combinations.
In tile, a label combination is for training when $Y_1 < 5$ or $Y_2 < 5$.
It is similar to the split for illustrative example.
In One-label, a label combination is for training when $Y_1 < 9$ or $Y_2 < 1$.
In such a case, only $(9, 0)$ contains $Y_2=0$.
It is similar to the case of one-shot learning.
The results for the feed-forward neural network are in Figure~\ref{fig:label_combination_results} and Table~\ref{tab:label_combination_results}.
It is similar to the results in the experiment section.

\begin{table}[!ht]
\centering
\caption{Accuracy (mean $\pm$ std \%).
Experimental results for other label combinations.}
\label{tab:label_combination_results}
\begin{tabular}{lrr|rr|rr}
& \multicolumn{2}{c|}{Test Sample Accuracy}
& \multicolumn{2}{c|}{Test Set Accuracy}
& \multicolumn{2}{c}{Random Set Accuracy} \\
& \multicolumn{1}{c}{Individual} & \multicolumn{1}{c|}{Shared} 
& \multicolumn{1}{c}{Individual} & \multicolumn{1}{c|}{Shared}
& \multicolumn{1}{c}{Individual} & \multicolumn{1}{c}{Shared} \\
\hline
Tile & 33.0 {\small$\pm$ \ \ 3.9\par} & 8.3 {\small$\pm$ 1.1\par} & 36.5 {\small$\pm$ \ \ 4.6\par} & 9.2 {\small$\pm$ 1.1\par} & 9.7 {\small$\pm$ 2.5\par} & 1.5 {\small$\pm$ 0.1\par} \\
Oneshot & 20.9 {\small$\pm$ 13.3\par} & 1.4 {\small$\pm$ 0.8\par} & 23.1 {\small$\pm$ 14.9\par} & 1.5 {\small$\pm$ 0.8\par} & 0.0 {\small$\pm$ 0.0\par} & 0.0 {\small$\pm$ 0.0\par}
\end{tabular}
\end{table}

\end{document}